\documentclass[10pt,twocolumn,letterpaper]{article}

\usepackage[pagenumbers]{cvpr} %

\usepackage[dvipsnames]{xcolor}

\usepackage{xcolor}
\definecolor{cvprblue}{rgb}{0.21,0.49,0.74}
\usepackage[pagebackref,breaklinks,colorlinks,citecolor=cvprblue]{hyperref}

\def\paperID{3394} %
\def\confName{CVPR}
\def\confYear{2024}

\newcommand{\ourdata}[0]{Beam-splitter Event Agile Human Motion Dataset}
\newcommand{\ourdatashort}[0]{BEAHM}
\newcommand{\ours}[0]{EvHuman}

\definecolor{claude_red}{RGB}{255,0,0}
\colorlet{lightgray}{gray!80}
\newcommand{\claude}[1]{{\color{claude_red} #1}}

\begin{document}
\title{Continuous-Time Human Motion Field from Events}

\newif\ifshowsupplementary
\showsupplementaryfalse

\newif\ifshowmainbib
\showmainbibfalse

\author{Ziyun Wang$^{1}$, Ruijun Zhang$^{2}$, Zi-Yan Liu$^{1}$, Yufu Wang$^{1}$,
Kostas Daniilidis$^{1,2}$\\
$^{1}$University of Pennsylvania, USA.\\
$^{2}$Archimedes, Athena RC
\\
}

\twocolumn[
\maketitle
\begin{center}
    \captionsetup{type=figure}
    \vspace{4pt}
    \includegraphics[clip, trim={1cm, 6.5cm, 1cm, 6.5cm}, width=.96\textwidth]{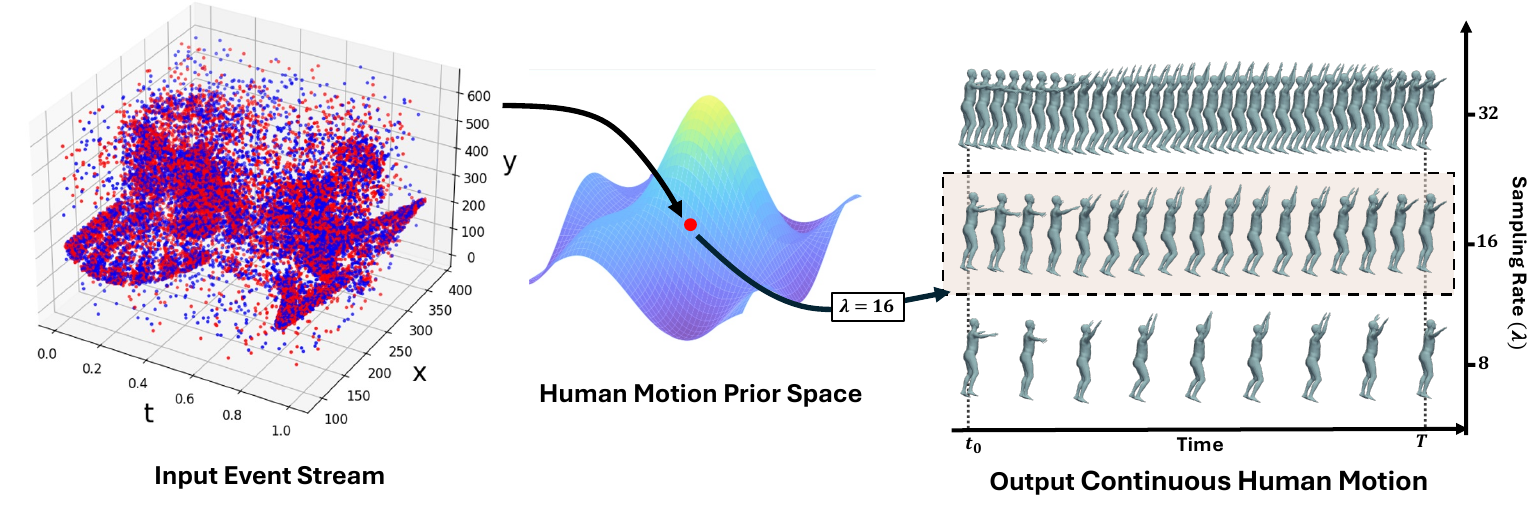}
    \vspace{0pt}
    \captionof{figure}{
        \textbf{\ours{}} predicts a set of global and local latent codes from an event stream to represent continuous-time human motions. The latent codes are decoded by a neural human motion prior in a time-continuous MLP network that can be queried at any time resolution in parallel efficiently. The human sequences on the rigth are decoded with a test event stream in MMHPSD~\cite{zou2021eventhpe}.
        }
    \label{fig:teaser}
\end{center}

]
\begin{abstract}
This paper addresses the challenges of estimating a continuous-time human motion field from a stream of events. Existing Human Mesh Recovery (HMR) methods rely predominantly on frame-based approaches, which are prone to aliasing and inaccuracies due to limited temporal resolution and motion blur. In this work, we predict a continuous-time human motion field directly from events, by leveraging a recurrent feed-forward neural network to predict human motion in the latent space of possible human motions. Prior state-of-the-art event-based methods rely on computationally intensive optimization across a fixed number of poses at high frame rates, which becomes prohibitively expensive as we increase the temporal resolution. In comparison, we present the first work that replaces traditional discrete-time predictions with a continuous human motion field represented as a time-implicit function, enabling parallel pose queries at arbitrary temporal resolutions. 
Despite the promises of event cameras, few benchmarks have tested the limit of high speed human motion estimation. We introduce \ourdata{}—a hardware-synchronized high-speed human dataset to fill this gap. On this new data, our method improves joint errors by 23.8 \% compared to previous event human methods, while reducing the computational time by 69\%.

\end{abstract}
    
\section{Introduction}
\label{sec:intro}

Human Mesh Recovery (HMR) methods recover the full 3D mesh of a moving human from a video, which has been a core research problem in computer vision. However, with highly dynamic human motions, performing such tasks with traditional cameras is challenging because frame-based cameras can only provide sampling of human motions at a limited frame rate.
First, it is challenging to predict the correct motion when the subject is moving fast and the time resolution of a video is low. Additionally, fast motions are often accompanied by motion blur that squashes motion information over time, preventing the network from obtaining the correct pose information.

To address these issues, researchers have begun exploring event cameras as an alternative sensor modality~\cite{xu2020eventcap, zou2021eventhpe, zhu2021eventgan, calabrese2019dhp19, zou2023event}. Event cameras are known for their high temporal resolution, high dynamic range, and low data throughput. Due to their asynchronous design, there is no fixed global shutter time, which helps mitigate motion blur. Event data captured with a static camera are inherently background and aliasing free, providing a continuous motion signal rather than discretized frames. Additionally, event cameras can estimate the motion of 2D pixels more robustly because only changes in the scene are recorded~\cite{zhu2018ev, zhu2019unsupervised, gehrig2021raft, hamann2024motion}. These advantages make event cameras ideal sensors for capturing high-speed human motion under various lighting conditions. Despite the higher temporal resolution and better motion signals, existing approaches assume that the predicted poses are represented as a sequence of discrete poses, which is computationally expensive to optimize and requires a fixed number of predicted poses known a priori. For high-speed prediction, the optimization performance can be up to 6750 times slower than real time~\cite{xu2020eventcap}. Although learning-based methods are faster~\cite{zou2021eventhpe, calabrese2019dhp19}, the inference time scales linearly with the number of query poses, and the pose error increases due to chaining short predictions.

In this work, we introduce \textit{\ours{}}, the first learning-based HMR approach that directly outputs a continuous-time motion field from events, enabling the query of human pose at any arbitrary timestamp within the event stream. Our approach significantly outperforms the prior methods across a variety of HMR metrics while significantly reducing the computational time of prediction sequences of human poses at high temporal resolutions.

Unlike existing methods that predict human poses frame-by-frame, \ours{} learns latent codes of human motion, which are then decoded with a human motion prior network pretrained on a wide range of diverse human motions. The decoder itself is a time-continuous function,
predicting both root and local poses for any specified query time. A global motion predictor takes in the predicted poses, joint positions, and velocities, and maps them to global velocities.
Our training process incorporates traditional supervised losses and introduces a novel event-based contrast maximization loss using vertex optical flow derived from the predicted human meshes.
During inference, unlike optimization methods like EventCap~\cite{xu2020eventcap}, \ours{} does not need a fixed set of initial guessed poses. Instead, it encodes an entire event stream once and can predict pose prediction at any time resolution by evaluating at arbitrary timestamps.

We evaluated our method against state-of-the-art event and image human methods on MMHPSD~\cite{zou2021eventhpe} and our novel \textbf{\ourdata{} (\ourdatashort{})}. \ourdatashort{} was collected with a custom-built event/image beam splitter and multiple high frame-rate cameras to capture high-speed human mesh labels. Precise hardware synchronization aligns events and images temporally for accurate benchmarking and ground-truth labeling. We make publicly available all raw events, images, the beam splitter design, and data collection software. Our main contributions are as follows.
\begin{itemize}
    \item We introduce the first feed-forward event-based continuous-time human motion field leveraging neural human motion priors, advancing the state of the art performance for event-based human mesh by 23.8 \% while reducing the computational time by 69 \%.
    \item We design a novel event-based human mesh motion loss that explicitly maximizes event contrast based on flows rendered from our continuous-time human motion field.
    \item We collected a new high-resolution event-based human pose dataset, \ourdata{}, that provides ground truth meshes at 120 FPS.
\end{itemize}

\section{Related Work}
\label{sec:related}

\textbf{Event-based Human Pose Estimation}
3D human pose estimation is grouped into two categories. The first category estimates 3D skeletal joint positions~\cite{pavlakos2017coarse, Sun_2017_ICCV, Pavlakos_2018_CVPR, Moreno-Noguer_2017_CVPR, Tekin_2017_ICCV, Martinez_2017_ICCV}. The second category, which is more related to our problem, recovers a parametric 3D human mesh, such as the SMPL model~\cite{10.1145/3596711.3596800}. To recover SMPL parameters, methods employ an optimization-based approach by fitting to the image evidence~\cite{opt_smplify, opt_video, opt_multiview}, or learn from the data to directly regress the pose and shape parameters~\cite{hmr, goel2023humans, spin, pare, bev, camerahmr}. Our method is a regression approach. However, instead of directly regressing the parameters, we predict a latent representation~\cite{he2022nemf} that is decoded to the SMPL parameters. Recovery of global human motion from a dynamic camera is more challenging and often requires additional sensors~\cite{Kaufmann_2023_ICCV, 10342291, Marcard_2018_ECCV} or integration with SLAM techniques~\cite{smartmocap, egobody, slahmr, kocabas2024pace, wang2024tram}. In this study, we assume a static camera setup; however, event-based approaches face difficulties with static humans because no events are generated.

Event-based human pose estimation has advanced through datasets like DHP19 \cite{calabrese2019dhp19}, which support 2D joint detection and triangulation. Recent developments include TORE’s volume-based representation for joint lifting \cite{baldwin2022time}, Scarpellini’s end-to-end single-camera framework \cite{scarpellini2021lifting}, and Chen’s point aggregation approach \cite{chen2022efficient}. For 3D mesh recovery, EventCap \cite{xu2020eventcap} optimizes human mesh through tracking joints through events, while EventHPE \cite{zou2021eventhpe} learns 3D human pose through poses and optical flow supervision. A spiking-based extension is used to improve energy efficiency for event-based HPE~\cite{zou2023event}.

\textbf{Unsupervised Event Optical Flow Estimation}
Learning-based flow estimation from events has been extensively studied in recent years~\cite{zhu2018ev, zhu2019unsupervised, gehrig2021raft, ye2018unsupervised, bardow2016simultaneous, pansingle, low2020sofea, zhang2022formulating, wang2025evimo, wang2023event}. Contrast Maximization (CM) methods have demonstrated competitive performance in optical flow estimation using only event data~\cite{ye2018unsupervised, zhu2018realtime, gallego2018unifying, stoffregen2019event, hamann2025motion}. 
Ye et al.~\cite{ye2018unsupervised} introduced a pipeline for learning Egomotion, which is guided by aligning adjacent event slices using predicted rigid flow. Zhu et al.~\cite{zhu2019unsupervised} introduced a novel timestamp-based motion loss to enhance the robustness of contrast calculations. Gallego et al.~\cite{gallego2018unifying} introduced a comprehensive framework, which extends the application of contrast maximization to both flow and depth estimation. Gallego et al.~\cite{gallego2019focus} examined multiple forms of contrast maximization functions, offering an extensive empirical comparison between various loss functions. A key advantage of the contrast maximization approach is that it requires only event data as input, enabling fine motion supervision even when ground truth is unavailable.

\vspace{-.4cm}

\paragraph{Learned Human Motion Priors}
Various techniques have been proposed to provide priors for motion estimation~\cite{ling2020character, li2002motion, rose1998verbs, holden2017phase, komura2017recurrent, ling2020character, liu2005learning, brand2000style}. Unlike physics-based methods, these approaches learn probabilistic transitions between states from motion capture data~\cite{mahmood2019amass}. Motion variational auto-encoders can be trained to animate single characters by sampling the distribution of possible motions~\cite{ling2020character}. HuMoR~\cite{Rempe_2021_ICCV} learns a 3D human dynamical model based on the conditional variational autoencoder, which describes the transition probability between two consecutive human states. He et al.~\cite{he2022nemf} employ an implicit function to represent continuous human motion. PACE~\cite{kocabas2024pace} extends this method and shows superior performance in world-grounded human motion estimation. Unlike NeMF-based optimization, which fits to a fixed number of initialized poses, our approach takes full advantage of the  continuous poses in training, by computing motion induced optical flow to self-supervise event networks.

\section{Method}
\label{sec:method}

\begin{figure*}
    \centering
    \includegraphics[trim=4.6cm 7cm 4.8cm 6cm,clip,width=0.985\linewidth]{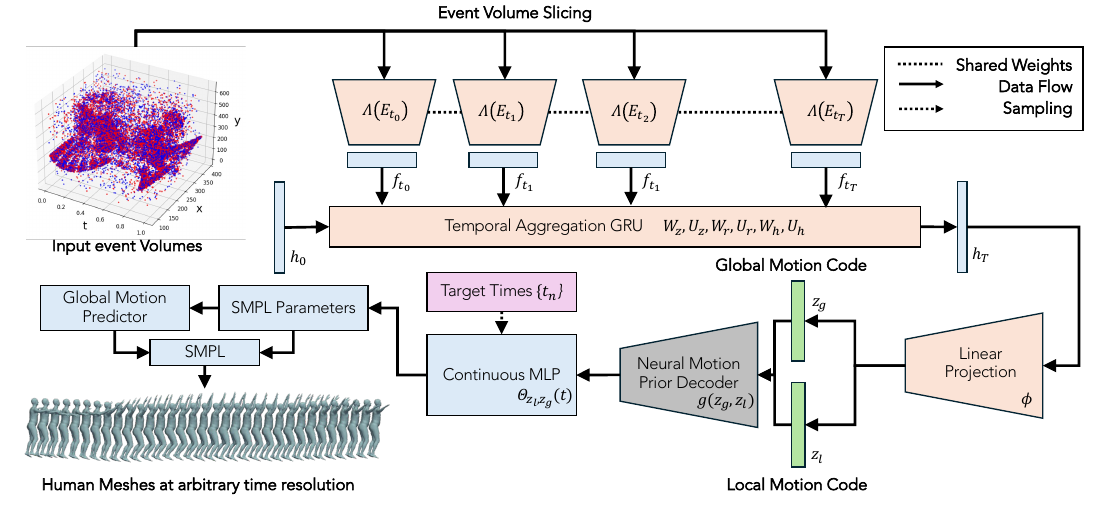}
    \caption{\textbf{Pipeline:} \ours{} takes in a continuous stream of events. The input event volumes are passed into a set of shared encoders. The encoded motion-rich features are further processed with a temporal aggregation network that iteratively refine a hidden state $h_0$. We apply two linear projects to project the terminal hidden state into the global and motion latent code. The latent codes are decoded with a pre-trained neural motion prior decoder to a MLP network that predicts SMPL parameters and global translation at any time.}
    \label{fig:pipeline}
\end{figure*}

Given an ordered set of events $E = \{x_i, y_i, t_i, p_i\}, t_i \in [0, T]$, from a monocular event camera, our objective is to predict the human mesh at any arbitrary time $t$. Following prior work~\cite{xu2020eventcap, zou2021eventhpe}, we assume the initial pose at the beginning of the sequence and the shape parameters are known.

We parameterize the poses as a continuous-time function $f_{E}(t)$ that maps time $t$ to a set of SMPL parameters 
$\{\Gamma_{t}\}, \{\tau_t\}, \{\gamma_{t}\}$. Here $\Gamma_t$ denotes the local body poses, $\gamma_{t}$ the root rotation and $\tau_t$ the root translation. 
We describe the generative human motion model that we use for optimization based on the features of events (\textbf{\cref{sec:generative}}), and the human motion predictor from events (\textbf{\cref{sec:predictor}}). 
Then we discuss how we predict global poses from our decoded continuous-time local SMPL meshes (\textbf{\cref{sec:adaptation}}).
We then describe our novel event human contrast loss in which the predicted human motion are supervised using only input events (\textbf{\cref{sec:events}}). Finally, we describe the loss functions and our training strategy (\textbf{\cref{sec:training}}). An overview of the proposed pipeline can be found in~\cref{fig:pipeline}.
\subsection{Generative Human Motion Models\label{sec:generative}}
While humans can perform a variety of motions, these movements are inherently constrained by the physical limits of human joints and the natural distribution of common motion patterns. As a result, plausible human motions exist in a small subspace within the larger space of all poses. Understanding this prior distribution can help learning-based networks recover poses, especially in challenging scenarios such as occlusion, motion blur, and aliasing. This is particularly important for event-based data, which can be noisy and may miss information due to insufficient movement.

\textbf{Neural Human Motion Prior.} NeMF~\cite{he2022nemf} trained a variational auto encoder on large scale human pose data. This section focuses on searching for a point in the latent space given event data.
Given a pre-trained VAE decoder $p_{\phi} (\mathbf{x} | \mathbf{z})$, where $\mathbf{z}$ is a latent code and $\mathbf{x}$ is the random variable representing human motions, the problem of finding a good $\mathbf{z}$ that fits our observed motion $\mathbf{x}^*$  can be thought of as maximizing the posterior of $p(\mathbf{z} | \mathbf{x}^*)$. Using the Bayes' rule, the log likelihood of the posterior is:
\begin{align}
\log p(\mathbf{z}|\mathbf{x}^*) &= \log p(\mathbf{x}^* | \mathbf{z}) + \log p(\mathbf{z}) - \log p(\mathbf{x}^*)
\end{align}
Since $p(\mathbf{x}^*)$ is constant with respect to $z$, it is irrelevant in optimization. The optimization objective becomes:
\begin{align}
    \mathbf{z} = \arg\max_\mathbf{z} \log p(\mathbf{x}^* | \mathbf{z}) + \log p(\mathbf{z}).
\end{align}

Here, maximizing $p(\mathbf{x}^* | \mathbf{z})$ is equivalent to minimizing the reconstruction loss between the predicted human mesh and the ground truth. In practice, we compute this loss as the weighted sum of several losses of the joint rotations, 3D and 2D keypoint locations, described in~\cref{sec:training}. In this work, we use a neural network $g$ that predicts the latent code $z$ from events $E$, and optimize the parameters $\theta$ of $g(\cdot)$.
\begin{align}
   \theta =  \arg\max_\mathbf{\theta} \sum_i^D  \log p(\mathbf{x}^*_i | g(E_i; \theta)).
\end{align}
The decoder of the latent space can be parameterized by a Multi Layer Perceptron (MLP) that takes a pair of latent code and outputs $\Gamma_{t}, \gamma_{t}$, which are parameters used by SMPL to recover the full mesh shape:
\begin{align}
    \Theta_{\mathbf{z}_l, \mathbf{z}_g}(t): t \rightarrow (\Gamma_{t}, \gamma_{t})
\end{align}
Given latent codes $\mathbf{z_l}$ and $\mathbf{z_g}$ predicted from an event-based model encoder, we can sample human poses at timestamps $\{t_i\}$ from $\Theta_{\mathbf{z}_l, \mathbf{z}_g}(t)$ directly in parallel. 

\subsection{Event Human Motion Predictor\label{sec:predictor}} 
The Event Human Motion Predictor predicts the latent codes $\mathbf{z_l}, \mathbf{z_g}$ from the input events. An important difference between events and images is that events do not exist for non-moving parts of the scene, assuming constant lighting. Thus, it is crucial to consider the temporal relationship among the stream of events, rather than predicting the pose independently for each event frames as in~\cite{calabrese2019dhp19}. 
We employ a Gated Recurrent Unit Network for this purpose. 
First, we extract features $\{\mathbf{f}_{t_0}, \mathbf{f}_{t_1} \cdots \mathbf{f}_{t_T}\}$ from sampled event volumes at time $\{t_0, t_1, \cdots t_T\}$ using a shared feature encoder $\Lambda$. We first initialize the hidden state $\mathbf{h_0}$ as a zero vector. Then we apply a Gated Recurrent Unit to aggregate these features temporally:
\begin{align}
    \mathbf{z}_t &= \sigma(\mathbf{W}_z \mathbf{f}_t + \mathbf{U}_z \mathbf{h}_{t-1} + \mathbf{b}_z) \\
    \mathbf{r}_t &= \sigma(\mathbf{W}_r \mathbf{f}_t + \mathbf{U}_r \mathbf{h}_{t-1} + \mathbf{b}_r) \\
    \tilde{\mathbf{h}}_t &= \tanh(\mathbf{W}_h \mathbf{f}_t + \mathbf{U}_h (\mathbf{r}_t \odot \mathbf{h}_{t-1}) + \mathbf{b}_h) \\
    \mathbf{h}_t &= (1 - \mathbf{z}_t) \odot \mathbf{h}_{t-1} + \mathbf{z}_t \odot \tilde{\mathbf{h}}_t
\end{align}

where $\mathbf{W}_z$, $\mathbf{U}_z$, $\mathbf{W}_r$, $\mathbf{U}_r$, $\mathbf{W}_h$, and $\mathbf{U}_h$ are learnable weight matrices, $\mathbf{b}_z$, $\mathbf{b}_r$, and $\mathbf{b}_h$ are biases, and $\sigma$ denotes the sigmoid activation function. At the final time step $T$, the hidden state \(\mathbf{h}_T\) encodes the accumulated temporal information. We then decode the target entity directly from \(\mathbf{h}_T\), leveraging it as a summary of the input sequence’s temporal features. In the end, we apply two separate linear projections $\phi$ to project from hidden state to the local latent codes $[\mathbf{z}_l, \mathbf{z}_g ] = \phi(\mathbf{h}_T) $, where $\phi$ is a learnable linear layer.

\begin{figure*}[h]
    \centering
    \includegraphics[clip, trim={1.4cm, 6.5cm, 1.1cm, 7cm}, width=\linewidth]{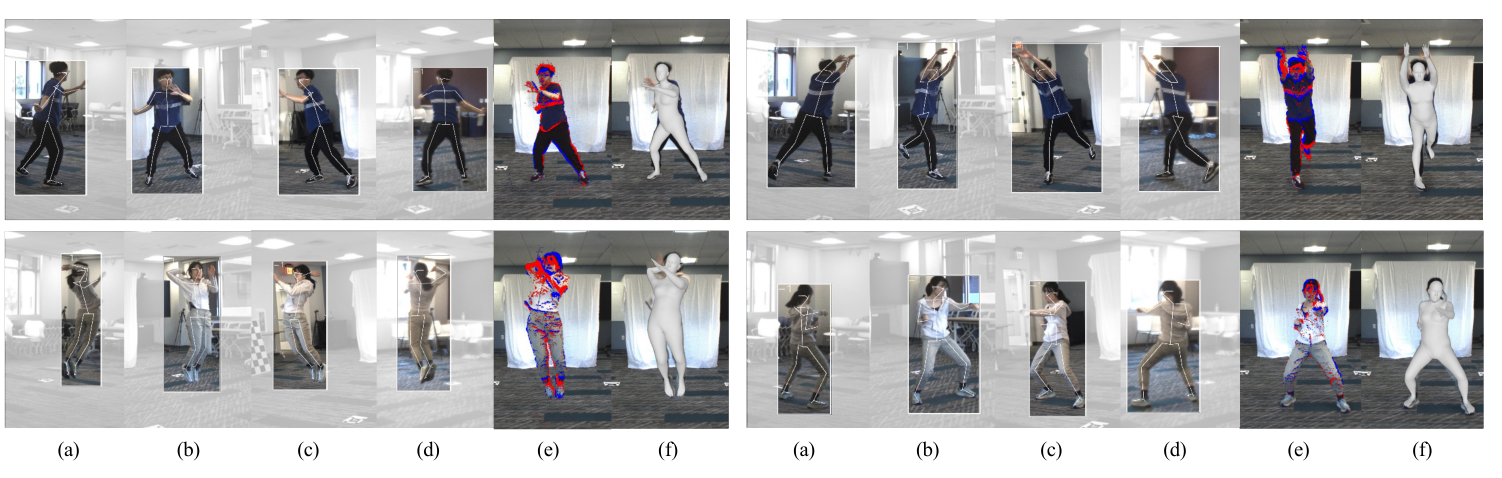}
    \caption{We present four example sequences from data collection of~\ours{}. Each sequence, from left to right, includes: \textbf{(a-d)} Four multi-camera images with bounding boxes and skeleton estimations via EasyMocap~\cite{easymocap}. \textbf{(e)} Events displayed on the beam splitter RGB camera. \textbf{(f)} The estimated mesh model superimposed on the beam splitter RGB camera.}
    \label{fig:dataset}
\end{figure*}
\begin{figure}[ht]
    \centering
    \includegraphics[clip, width=0.7\linewidth, trim={4.3cm 6.5cm 11cm 9cm}]{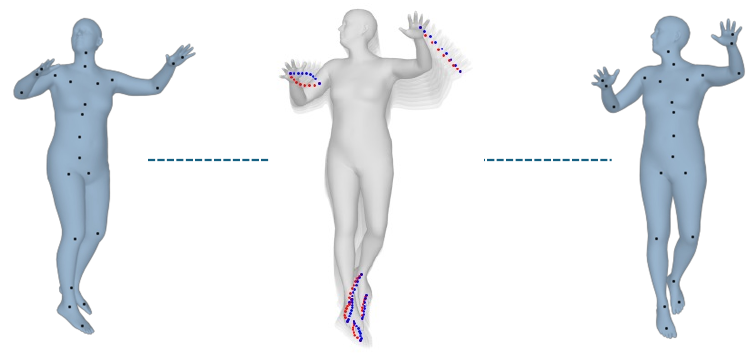}
    \caption{Continuous-time decoding compared to Interpolation. Left and right: Start and end pose. Middle: full joint trajectory. Interpolated key points marked in red and continuous pose in blue. }
    \label{fig:inter}
\end{figure}

\subsection{Global Motion Estimation}
Assuming a known initial rotation $R_i(t_0)$ at the beginning of the sequence and the decoded rotations $\{\hat{R}_i(t)\} \forall t \in [0, T]$, the relative rotations with respect to the initial pose is given as
$\Delta R_i(t) = \hat{R}^{-1}_i(t_0) \hat{R}_i(t)$
and adapt the motion to the known initial pose:
    $\bar{R_i}(t) = R_i (t_0)\Delta R_i(t)$.

\textbf{Global Motion Prediction (GMP)} We model the global translation of the subject as a function of the local pose, following prior work in global human pose estimation~\cite{li2021task, zhou2020generative}. Given a decoded local pose from our Event Human Motion Predictor, we estimate the velocity $\dot{\tau}_t$ at time $t$ rather than direct translation $\tau_t$. Following~\cite{kocabas2024pace, he2022nemf}, the root velocities, root height, and contacts are predicted from a convolutional neural network from joint rotations and velocities. We denote the global translation function $\mathcal{P}(\{\bar{R}_i(t)\}, \{\omega_i(t)\}, \{P_i(t)\} , \{\dot{P}_i(t)\})$, which maps the joint rotations $\{\bar{R}_i(t)\}$, angular velocities $\{\omega_i(t)\}$, joint positions $\{\hat{P}_i(t)\}$ and joint velocities $\{\dot{P}_i(t)\}$ to the relative root velocities $\dot{\tau}_t$. Finally, the velocity is integrated using Euler's method iteratively forward $\tau_{t + \delta t} = \tau_t + \dot{\tau}_t \delta t$. Please see details of GMP in the Supplementary Material.
\label{sec:adaptation}

\subsection{Human Mesh Event Contrast Maximization}

\begin{figure}[ht]
    \centering
    \includegraphics[clip, width=0.9\linewidth, trim={0.5cm 9cm 18.35cm 5.8cm}]{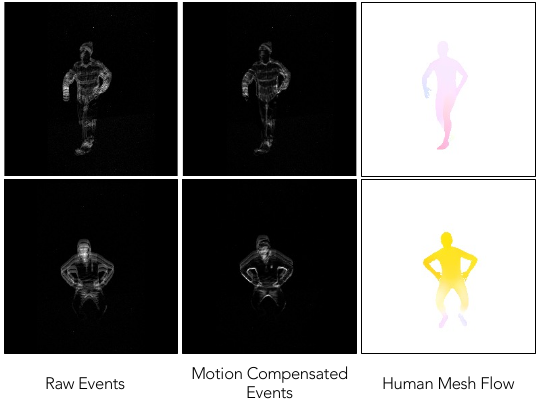}
    \caption{\label{fig:contrast}Human Event Contrast Maximization. \textbf{Left}: Raw event IWE. \textbf{Middle}: Motion-compensated events using estimated human motion. \textbf{Right:} Dense motion field from our continuous-time human motion field. Color indicates direction of optical flow.} 
    \label{fig:computation}
\end{figure}

\label{sec:events}
Given a continuous motion model $\Theta_{\mathbf{z}_l, \mathbf{z}_g}(t)$ and the translation $\tau_t$ computed in~\cref{sec:adaptation}, the vertices can obtained by:
\begin{align}
    V^v_t = \mathcal{W} (\mathcal{S} (\Theta_{\mathbf{z}_l, \mathbf{z}_g}(t), \beta, \tau_t), v)
\end{align}
where $\mathcal{S}$ is the parametric SMPL~\cite{loper2023smpl} that returns a global human mesh parameterized by joint poses and shape parameter $\beta$, $\mathcal{W}$ is a skinning function on the mesh, and $v$ the vertex index. This continuous-time motion model gives us the full trajectory of vertices in 3D. The motion field defined between two times $t_i$ and $t_j$ can be computed by subtracting the 2D location of the same vertex: 
\begin{align}
    \mathbf{F}^{shape}_{i,j,v} = \mathbf{1}_{vis}(v) (\pi(V^v_{t_i}) - \pi(V^v_{t_j})),
\end{align}
where $\pi$ is the perspective projection function given known intrinsics. While it is easy to use all vertices in this loss, the back of the human motion (with respect to the camera) can produce erroneous flow. We use differentiable renderers to render optical flow on mesh triangles using Barycentric coordinates~\cite{Laine2020diffrast} so that flow can be differentiable with respect to the SMPL parameters. We denote the visibility of the vertices as $\mathbf{1}_{vis}(v)$ based on mesh rasterization.
Inspired by unsupervised event-based optical flow methods~\cite{gallego2019focus, zhu2019unsupervised, hamann2024motion, ye2018unsupervised}, we maximize the variance of the Image of Warped Events (IWE). Intuitively, this means that the events will look sharp if we properly compensate them by the correct motion. 
Flow from low-parameter models such as SMPL is uniquely compatible with Contrast Maximization because it avoids problems such as event collapse. 
Qualitative results can be found in \cref{fig:contrast}. Given the events $E_{ij} = \{\mathbf{x_k}, p_k, t_k\}$ between $t_i$ and $t_j$, the motion-compensated events become
\begin{align}
    \mathbf{x_k'} &= \mathbf{x_k} - \mathbf{F}^{shape}_{i,j,v}(\mathbf{x_k}) (t_r - t_k).
\end{align}
The image of warped events (IWE) is defined as
\begin{align}
    I(\mathbf{x}; \mathbf{F}^{shape}_{i,j,v}) = \sum^{|E_{ij}|}_k b_k \delta(\mathbf{x} - \mathbf{x_k'}),
\end{align}
where $b_k$ is the polarity indicator that separates events into a positive and a negative image, and $\delta$ is the bilinear kernel function.
Various objective functions are described in detail~\cite{gallego2019focus}. We maximize the image variance:
\begin{align}
\text{Var}(I) &= \frac{1}{MN} \sum_{u=1}^{M} \sum_{v=1}^{N} \left( I(u, v) - \mu \right)^2 \\
\mu &= \frac{1}{MN} \sum_{u=1}^{M} \sum_{v=1}^{N} I(i, j).
\end{align}
Since we maximize the variance, the loss function is the negative variance $\mathcal{L}_c = - \text{Var}(I)$.

\subsection{Training\label{sec:training}}
\textbf{Input Representation.} We use event volumes~\cite{wang2022evac3d, zhu2018ev, wang2022ev} as the input representation. For a stream of events $E = \{x_i, y_i, t_i, p_i\}$, an event volume is computed as 
\begin{align}
E(x,y,t)=&\sum_{i} p_i k_b(x-x_i)k_b(y-y_i)k_b(t-t^*_i)\label{eq:event_volume},
\end{align}
where events are inserted using bilinear kernel $k_b$ into a voxel. Event voxels preserve motion information and can be used with standard grid-based encoder architectures.

\textbf{Loss Functions.} We compute the geodesic loss from the predicted joint rotations $\{\hat{R_t}\}$ and the ground truth $\{R_t\}$:
\begin{align}
    \quad \mathcal{L}_{\text{ori}} = \sum_{t=1}^{T} || \log(R_t \hat{R}^T_t) ||_F.
\end{align}
The loss between the predicted translations $\{\hat{T}_t\}$ and ground truth translation $\{T_t\}$ in the camera frame:
\begin{align}
    \mathcal{L}_t = \sum^T_{t=1} ||T_t - \hat{T}_t||_2^2.
\end{align}
We compute key point loss between predicted 3D joint positions $\{\hat{P}^k_t\}$ and ground truth 3D joint positions $\{P^k_t\}$:
\begin{align}
    \mathcal{L}_{3D} = \sum^T_{t=1} \sum^{24}_{k=1} ||P^k_t - \hat{P}^k_t||_2^2,
\end{align}
and between 2D prediction $\{\pi(\hat{P}^k_t)\}$ and ground truth $\{p^k_t\}$:
\begin{align}
    \mathcal{L}_{2D} = \sum^T_{t=1} \sum^{24}_{k=1} ||p^k_t - \pi(\hat{P}^k_t)||_2^2,
\end{align}
where $\pi$ is the perspective projection function.
We additionally follow EventHPE~\cite{zou2021eventhpe} to compute the cosine difference between the vertex flow computed from the predicted human motion field $F^{shape}_{t_i, t_j}$ and the event-based flow $F^{e}_{t_i, t_j}$ when a pre-trained flow network is available:
\begin{align}
\mathcal{L}_{flow} = \sum_{t=1}^{T} \sum_{v} \frac{<\mathbf{F}^{shape}_{t, v}
\mathbf{F}^{e}_{t, v}>}{||\mathbf{F}^{shape}_{t, v}||_2 \cdot ||\mathbf{F}^{e}_{t, v}||_2}
\end{align}
The flow network is trained in an unsupervised fashion using paired images and events, following Ev-FlowNet~\cite{zhu2018ev}.
Finally, we compute the unsupervised flow loss $\mathcal{L}_c$ as described in Sec.~\ref{sec:events}.
The final loss function is the weighted sum of the terms described above:
\begin{align}
    \mathcal{L} = &\lambda_{ori} \mathcal{L}_{ori} + \lambda_{t} \mathcal{L}_{t} + \lambda_{3D} \mathcal{L}_{3D} + \nonumber\\  
    &\lambda_{2D} \mathcal{L}_{2D} + \lambda_{flow} \mathcal{L}_{flow} + \lambda_{c} \mathcal{L}_{c}.
\end{align}
\textbf{Training Strategy.} We first train a global motion predictor (GMP) (\cref{sec:adaptation}) using ground-truth local motions as input. Then, we freeze the global motion predictor and train the event human motion predictor (\cref{sec:predictor}). In the end, we freeze the local motion prediction and fine-tune the GMP for 1 epoch. Details of training can be found in the Supplementary Material.

\section{\ourdata{} (\ourdatashort{})\label{sec:data}}

A key challenge in studying continuous event-based HMR is the lack of high-speed labeled datasets. We need high-speed labels of the human mesh, a diverse set of motions, and precise synchronization between events and ground truth. To address this, we collect \ourdata{} (\ourdatashort{}) compared with prior datasets~\cref{tab:data}. We used a single-objective beam splitter with a shared lens to obtain aligned events and images. The ground truth is obtained through four calibrated RGB cameras using a state-of-the-art multi-view human reconstruction technique~\cite{easymocap}.  To ensure precise timing, we built a custom trigger board to synchronize event camera and 120 FPS RGB cameras. We designed 40 diverse motions from slow walking to fast Karate kicking, covering a wide spectrum of difficulties.
\begin{table}[ht]
\centering
\caption{\label{tab:data}Comparison of event human pose estimation datasets.
}
\small{
\begin{tabular}{lccccccc}
\toprule
\textbf{Dataset}  & \textbf{Label} & \textbf{Label FPS} & \textbf{Sync} & \textbf{Motions} \\
\midrule
DHP19~\cite{calabrese2019dhp19}       &2D joints & 100    & Hard  & 33\\
EventCap~\cite{xu2020eventcap}     &Mesh   & 100   & -     & 12       \\ 
MMHPSD~\cite{zou2021eventhpe}       &Mesh   & 15    & Soft  & 21\\
CDEHP~\cite{shao2024temporal}        &2D joints & 60    & -  & 25\\
\ourdatashort{}         &\textbf{Mesh}  & \textbf{120}   & \textbf{Hard}  & \textbf{40} \\
\bottomrule
\end{tabular}
}
\end{table}

\textbf{High speed labels}. Fast pose ground truth is critical for rapid human movements. Simple interpolation of joint rotations and translations may work for slow movements but fails for accelerated motions. \Cref{fig:dataset} illustrates this, with 120 FPS ground truth joints in red and the Slerp-interpolated joints in blue, showing notable differences in fast motion. Our analysis shows an average error as high as 25 mm, which shows the need for high-speed labeling.

\section{Experiments}
\label{sec:exp}

\begin{table*}[ht]
\centering
\caption{\label{tab:quant}Quantitative comparison on MMHPSD~\cite{zou2021eventhpe} and our \ourdatashort{}. DHP19$^{\dag}$ uses the groundtruth depth for each joint. We include their upper-bound performance but exclude their performance from ranking. Data marked with ${}^\ast$ is sourced from the original papers.}
\begin{tabular}{lcccccc}
\toprule
\textbf{Models} & \textbf{MPJPE $\downarrow$} & \textbf{PA-MPJPE $\downarrow$} & \textbf{PEL-MPJPE $\downarrow$} & \textbf{PCKh@0.5 $\uparrow$} & \textbf{Input Modality} \\
\midrule
\multicolumn{6}{c}{\textbf{MMHPSD}}\\
\midrule
${}^\dag$DHP19 & ${}^\dag${72.42} & ${}^\dag${65.87} & ${}^\dag${74.04} & ${}^\dag${0.81} & Events \\
${}^\ast$HMR & - & 64.78 & 95.32 & 0.61 & Images \\
${}^\ast$EventCap (\textit{HMR Init}) & - & 62.62 & 89.95 & 0.64 & Images + Events \\
EventHPE (\textit{w/o HMR Feats}) & 81.06 & 45.86 & 58.90 & 0.84 & Events \\
${}^\ast$EventHPE &  71.79 & 43.90 & 54.96 & 0.85 & Images + Events\\
Ours & \textbf{67.66} & \textbf{39.16} & \textbf{52.23} & \textbf{0.86} & Events \\
\midrule
\multicolumn{6}{c}{\textbf{\ourdatashort{}}}\\
\midrule
${}^\dag$DHP19 & ${}^\dag${46.75} & ${}^\dag${42.39}  & ${}^\dag${48.15}  & ${}^\dag${0.86}  & Events\\
HMR & - &68.73  &112.55  &0.42  & Images\\
HMR 2.0 & - & 52.12 & 80.19 &0.60 & Images\\
EventCap (w/ HMR Init) & - & 65.81 & 93.94 & 0.70 & Images + Events\\
EventHPE (\textit{w/o HMR Feats}) & 65.28 & 38.91 & 52.31 & 0.86 & Events  \\
Ours  \textit{w/o Fine-tune} & 67.01 &  39.16 &  52.23  &  0.86 & Events\\
Ours  \textit{w/o} $\mathcal{L}_c$ & 50.77 & 30.22 &  41.26& 0.91 & Events  \\
Ours   & \textbf{49.74} & \textbf{30.05} &  \textbf{41.06} &  \textbf{0.92} & Events\\

\bottomrule
\end{tabular}
\end{table*}

\subsection{Datasets and Metrics}
\textbf{MMHPSD}~\cite{zou2021eventhpe} is a recent event-based human dataset that uses a multi-camera setup. It uses a CeleX-V event sensor that captures synchronized events and grayscale images. The main sensors and software-synchronized with ground truth sensors. The ground truth is provided on each frame at 15 FPS. 

\textbf{\ourdatashort{}.} The detailed description of our data is provided in~\cref{sec:data}. We evaluate on a similar time duration of approximately 1 second, which corresponds to skipping 8 labeled frames at 120 FPS, with 8 evaluation points in each window. In addition, we provide results of 120 FPS evaluation in \cref{tab:120fps}. We report 3D human joint metrics including MPJPE, Pelvis Adjusted MPJPE, Procrustes-Aligned MPJPE and PCKh@0.5. Pelvis Adjusted MPJPE eliminates the root transformation, which is commonly used with centered human crops. We report unadjusted MPJPE to include global translation in our evaluation.

\subsection{Baseline Methods}
We compare with several image and event human baselines. Similar to EventHPE~\cite{zou2021eventhpe}, we compare with \textbf{HMR} and \textbf{HMR 2.0}, both image-based methods. We compute predictions on images with HMR and HMR 2.0 and only interpolate if the ground truth rate is higher than the image frame rate as in~\cref{tab:120fps}. We re-implemented \textbf{EventCap}~\cite{xu2020eventcap} for comparison. 
In additon to the original EventHPE, we provide an additional baseline by retraining the \textbf{EventHPE} network without HMR features, denoted as \textit{EventHPE (w/o HMR Feats)}. \textbf{DHP19}~\cite{calabrese2019dhp19} is a 2D keypoint method. We re-trained DHP19 based on the SMPL's 24 joints and lifted their 2D joints into 3D using ground-truth depth, providing their 3D performance upper bound.

\begin{figure*}[ht]
    \centering
    \includegraphics[clip, trim={2.5cm, 5.7cm, 2.6cm, 5.6cm}, width=0.97\linewidth]{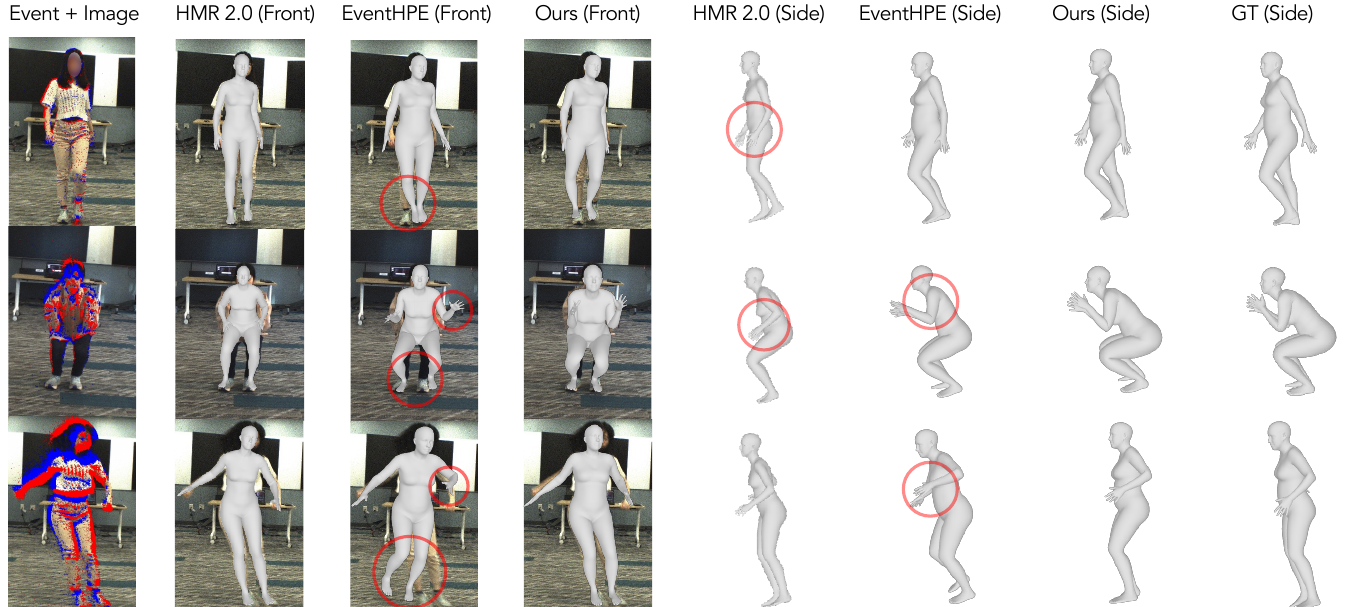}
    \caption{\textbf{Qualitative results of comparison with baseline methods}. We mark the erroneous predictions with red circles. For each method, we include the front view and the side view. In the first row, HMR 2.0 predicts erroneous unnatural motion seen from the side. The bottom two rows show the robustness of our method in fast motions compared to baseline methods. }
    \label{fig:our_qual}
    \vspace{-.5cm}
\end{figure*}

\subsection{Comparisons}
\textbf{Comparison on MMHPSD~\cite{zou2021eventhpe}} The quantitative and qualitative comparisons on MMHPSD dataset are presented separately in \cref{tab:quant} and \cref{fig:our_qual} separately. As evident in \cref{tab:quant}, \ours{} outperforms \textit{EventHPE (w/o HMR Feats)}, the best event-based baseline method, by 6.70 mm in PA-MPJPE and 6.67 mm in PEL-MPJPE. Our method beats the best overall method \textit{EventHPE} by 4.13 mm, 4.74 mm, 2.73 mm separately in MPJPE, PA-MPJPE, and PEL-MPJPE.

\textbf{Comparison on \ourdatashort{}}
Quantitative comparisons between \ours{} and baseline methods at a 15 FPS image frame rate, similar to EventHPE, are presented in \cref{tab:quant}. As shown in \cref{tab:quant}, our method improves MPJPE by 15.54 mm, PA-MPJPE by 8.86 mm, PEL-MPJPE by 11.25 mm and 0.07 in PCKh@0.5, over EventHPE. \ours{} outperforms the best image baseline by 21.07 mm in PA-MPJPE and 39.13mm in PEL-MPJPE. HMR 2.0 produces an erroneous mesh due to the per-frame nature of the method, which is illustrated in~\cref{fig:our_qual}.
Leveraging the high-speed labels of \ourdatashort{}, we provide a comparison at 120 Hz in \cref{tab:120fps}.
Our method improves MPJPE by 17 mm, PA-MPJPE by 5.90 mm, PEL-MPJPE by 7.16 mm, and 0.03 in PCKh@0.5, over the best event method. During training, EventHPE showed instability in translation estimation. Our method outperforms the results interpolated with HMR 2.0 by PA-MPJPE by 26.14 mm, PEL-MPJPE by 52 mm and 0.41 in PCKh@0.5. Qualitative results are shown in~\cref{fig:our_qual}.

\textbf{Ablation Studies.} We performed two ablation studies on the proposed component on our \ourdatashort{} dataset in~\cref{tab:quant}.
\textit{w/o} $ \mathcal{L}_c$ refers to without the human mesh event contrast maximization in~\cref{sec:events}.  \textit{w/o Fine-tune} means we do not fine-tune the global motion predictor function (GMP), as described in~\cref{sec:training}. The global motion predictor fine-tuning at the end significantly boosts the global translation performance of the model, by improving the MPJPE by 17.27 mm, PA-MPJPE by 8.11 mm, PEL-MPJPE by 11.17 mm and 0.06 in PCKh@0.5. Contrast maximization improved the global and local MPJPE performance.

\begin{table}[ht]
\vspace{-.2cm}
\centering
\caption{\label{tab:120fps} Evaluation at 120 FPS on \ourdatashort{}. DHP19$^{\dag}$ uses the groundtruth depth for each joint. We include their upper-bound performance but exclude their performance from ranking.}
\footnotesize{

\setlength{\tabcolsep}{3pt} %
\begin{tabular}{lcccc}
\toprule
\textbf{Models} &\textbf{MPJPE}& \textbf{PA-MPJPE } & \textbf{PEL-MPJPE} & \textbf{PCKh@0.5} \\
\midrule
${}^\dag$DHP19~\cite{calabrese2019dhp19} & ${}^\dag${46.15} & ${}^\dag${41.34}  & ${}^\dag${47.23}  & ${}^\dag${0.86} \\
HMR 2.0~\cite{goel2023humans} & - & 56.23 & 93.08 & 0.51 \\
EventHPE~\cite{zou2021eventhpe} & 66.76 &35.97 & 48.24 &  0.89  \\
Ours & \textbf{49.76} & \textbf{30.07} & \textbf{41.08} & \textbf{0.92} \\
\bottomrule
\end{tabular}
\vspace{-.2cm}
}
\end{table}

\begin{figure}[ht]
    \centering
    \includegraphics[clip, trim={0cm, 0.4cm, 1cm, 0.6cm}, width=0.85\linewidth]{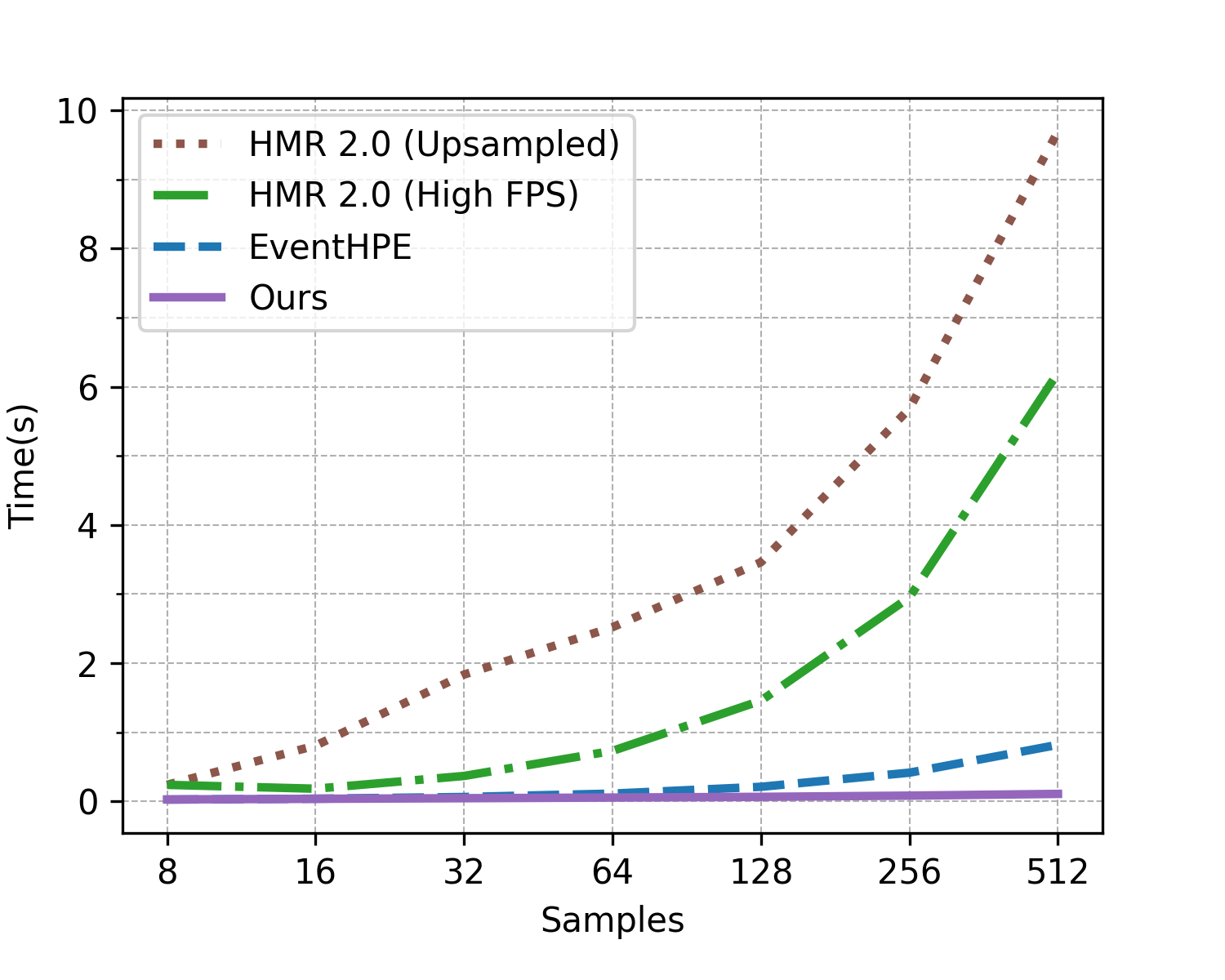}
    \caption{\label{fig:compute}\textbf{Computational Speed Comparison.} The computational time is plotted against the number of prediction frames. }
    \vspace{-.5cm}
\end{figure}

\subsection{Computational Speeds}
Due to the time continuity of our model, we can sample arbitrary number of poses in parallel. A computation time analysis is provided in~\cref{fig:compute}. For HMR 2.0, we present two settings: with video frame interpolation (\textit{HMR 2.0 + Upsampled)} and with raw high-speed input \textit{(HMR 2.0 + High FPS}). For \textit{HMR 2.0 + Upsample}, we upsample frames to the desired frame rate using FILM~\cite{reda2022film}. Our event baseline, EventHPE, predicts 8 frames at a time. As shown in \cref{fig:compute}, inference speeds are similar across methods at the training sampling rate (8 frames). However, the computational time for all three baseline methods increases significantly as the prediction length grows, due to the encoder processing larger input sequences. All evluations are conducted with an RTX 4080 deskptop GPU.

\vspace{-.2cm}
\section{Conclusion and Discussion}
This work introduces the first method that estimates a continuous human motion field from events. Leveraging a learned motion-prior latent space and an implicit motion decoder, our method allows for fast arallel inference at arbitrary temporal resolutions. The proposed method outperforms state-of-the-art event-based human pose methods while achieving 69 \% faster inference. We also contribute a hardware synchronized event-based human mesh dataset with high temporal resolution labeling, opening up valuable research opportunities in studying event-based human motion estimation. Despite its strengths, the method has limitations, including reliance on voxelized events and initial pose estimation. Future efforts will focus on reducing latency and eliminating the need for initialization.

\newpage
{
    \small
    \bibliographystyle{ieeenat_fullname}
    \bibliography{main}

\begin{thebibliography}{67}
\providecommand{\natexlab}[1]{#1}
\providecommand{\url}[1]{\texttt{#1}}
\expandafter\ifx\csname urlstyle\endcsname\relax
  \providecommand{\doi}[1]{doi: #1}\else
  \providecommand{\doi}{doi: \begingroup \urlstyle{rm}\Url}\fi

\bibitem[eas(2021)]{easymocap}
Easymocap - make human motion capture easier.
\newblock Github, 2021.

\bibitem[Arnab et~al.(2019)Arnab, Doersch, and Zisserman]{opt_video}
Anurag Arnab, Carl Doersch, and Andrew Zisserman.
\newblock Exploiting temporal context for 3d human pose estimation in the wild.
\newblock In \emph{Proceedings of the IEEE/CVF Conference on Computer Vision and Pattern Recognition}, pages 3395--3404, 2019.

\bibitem[Baldwin et~al.(2022)Baldwin, Liu, Almatrafi, Asari, and Hirakawa]{baldwin2022time}
R~Wes Baldwin, Ruixu Liu, Mohammed Almatrafi, Vijayan Asari, and Keigo Hirakawa.
\newblock Time-ordered recent event (tore) volumes for event cameras.
\newblock \emph{IEEE Transactions on Pattern Analysis and Machine Intelligence}, 45\penalty0 (2):\penalty0 2519--2532, 2022.

\bibitem[Bardow et~al.(2016)Bardow, Davison, and Leutenegger]{bardow2016simultaneous}
Patrick Bardow, Andrew~J Davison, and Stefan Leutenegger.
\newblock Simultaneous optical flow and intensity estimation from an event camera.
\newblock In \emph{Proceedings of the IEEE conference on computer vision and pattern recognition}, pages 884--892, 2016.

\bibitem[Bogo et~al.(2016)Bogo, Kanazawa, Lassner, Gehler, Romero, and Black]{opt_smplify}
Federica Bogo, Angjoo Kanazawa, Christoph Lassner, Peter Gehler, Javier Romero, and Michael~J Black.
\newblock Keep it smpl: Automatic estimation of 3d human pose and shape from a single image.
\newblock In \emph{Computer Vision--ECCV 2016: 14th European Conference, Amsterdam, The Netherlands, October 11-14, 2016, Proceedings, Part V 14}, pages 561--578. Springer, 2016.

\bibitem[Brand and Hertzmann(2000)]{brand2000style}
Matthew Brand and Aaron Hertzmann.
\newblock Style machines.
\newblock In \emph{Proceedings of the 27th annual conference on Computer graphics and interactive techniques}, pages 183--192, 2000.

\bibitem[Calabrese et~al.(2019)Calabrese, Taverni, Awai~Easthope, Skriabine, Corradi, Longinotti, Eng, and Delbruck]{calabrese2019dhp19}
Enrico Calabrese, Gemma Taverni, Christopher Awai~Easthope, Sophie Skriabine, Federico Corradi, Luca Longinotti, Kynan Eng, and Tobi Delbruck.
\newblock Dhp19: Dynamic vision sensor 3d human pose dataset.
\newblock In \emph{Proceedings of the IEEE/CVF conference on computer vision and pattern recognition workshops}, pages 0--0, 2019.

\bibitem[Chen et~al.(2022)Chen, Shi, Ye, Yang, Sun, and Wang]{chen2022efficient}
Jiaan Chen, Hao Shi, Yaozu Ye, Kailun Yang, Lei Sun, and Kaiwei Wang.
\newblock Efficient human pose estimation via 3d event point cloud.
\newblock In \emph{2022 International Conference on 3D Vision (3DV)}, pages 1--10. IEEE, 2022.

\bibitem[Dong et~al.(2019)Dong, Jiang, Huang, Bao, and Zhou]{opt_multiview}
Junting Dong, Wen Jiang, Qixing Huang, Hujun Bao, and Xiaowei Zhou.
\newblock Fast and robust multi-person 3d pose estimation from multiple views.
\newblock In \emph{Proceedings of the IEEE/CVF conference on computer vision and pattern recognition}, pages 7792--7801, 2019.

\bibitem[Gallego et~al.(2018)Gallego, Rebecq, and Scaramuzza]{gallego2018unifying}
Guillermo Gallego, Henri Rebecq, and Davide Scaramuzza.
\newblock A unifying contrast maximization framework for event cameras, with applications to motion, depth, and optical flow estimation.
\newblock In \emph{Proceedings of the IEEE conference on computer vision and pattern recognition}, pages 3867--3876, 2018.

\bibitem[Gallego et~al.(2019)Gallego, Gehrig, and Scaramuzza]{gallego2019focus}
Guillermo Gallego, Mathias Gehrig, and Davide Scaramuzza.
\newblock Focus is all you need: Loss functions for event-based vision.
\newblock In \emph{Proceedings of the IEEE/CVF Conference on Computer Vision and Pattern Recognition}, pages 12280--12289, 2019.

\bibitem[Gehrig et~al.(2021)Gehrig, Millh{\"a}usler, Gehrig, and Scaramuzza]{gehrig2021raft}
Mathias Gehrig, Mario Millh{\"a}usler, Daniel Gehrig, and Davide Scaramuzza.
\newblock E-raft: Dense optical flow from event cameras.
\newblock In \emph{2021 International Conference on 3D Vision (3DV)}, pages 197--206. IEEE, 2021.

\bibitem[Goel et~al.(2023)Goel, Pavlakos, Rajasegaran, Kanazawa, and Malik]{goel2023humans}
Shubham Goel, Georgios Pavlakos, Jathushan Rajasegaran, Angjoo Kanazawa, and Jitendra Malik.
\newblock Humans in 4d: Reconstructing and tracking humans with transformers.
\newblock In \emph{Proceedings of the IEEE/CVF International Conference on Computer Vision}, pages 14783--14794, 2023.

\bibitem[Hamann et~al.(2024)Hamann, Wang, Asmanis, Chaney, Gallego, and Daniilidis]{hamann2024motion}
Friedhelm Hamann, Ziyun Wang, Ioannis Asmanis, Kenneth Chaney, Guillermo Gallego, and Kostas Daniilidis.
\newblock Motion-prior contrast maximization for dense continuous-time motion estimation.
\newblock \emph{arXiv preprint arXiv:2407.10802}, 2024.

\bibitem[Hamann et~al.(2025)Hamann, Wang, Asmanis, Chaney, Gallego, and Daniilidis]{hamann2025motion}
Friedhelm Hamann, Ziyun Wang, Ioannis Asmanis, Kenneth Chaney, Guillermo Gallego, and Kostas Daniilidis.
\newblock Motion-prior contrast maximization for dense continuous-time motion estimation.
\newblock In \emph{European Conference on Computer Vision}, pages 18--37. Springer, 2025.

\bibitem[He et~al.(2022)He, Saito, Zachary, Rushmeier, and Zhou]{he2022nemf}
Chengan He, Jun Saito, James Zachary, Holly Rushmeier, and Yi Zhou.
\newblock Nemf: Neural motion fields for kinematic animation.
\newblock \emph{Advances in Neural Information Processing Systems}, 35:\penalty0 4244--4256, 2022.

\bibitem[Henning et~al.(2023)Henning, Choi, Schaefer, and Leutenegger]{10342291}
Dorian~F. Henning, Christopher Choi, Simon Schaefer, and Stefan Leutenegger.
\newblock Bodyslam++: Fast and tightly-coupled visual-inertial camera and human motion tracking.
\newblock In \emph{2023 IEEE/RSJ International Conference on Intelligent Robots and Systems (IROS)}, pages 3781--3788, 2023.

\bibitem[Holden et~al.(2017)Holden, Komura, and Saito]{holden2017phase}
Daniel Holden, Taku Komura, and Jun Saito.
\newblock Phase-functioned neural networks for character control.
\newblock \emph{ACM Transactions on Graphics (TOG)}, 36\penalty0 (4):\penalty0 1--13, 2017.

\bibitem[Kanazawa et~al.(2018)Kanazawa, Black, Jacobs, and Malik]{hmr}
Angjoo Kanazawa, Michael~J Black, David~W Jacobs, and Jitendra Malik.
\newblock End-to-end recovery of human shape and pose.
\newblock In \emph{Proceedings of the IEEE conference on computer vision and pattern recognition}, pages 7122--7131, 2018.

\bibitem[Kaufmann et~al.(2023)Kaufmann, Song, Guo, Shen, Jiang, Tang, Z\'arate, and Hilliges]{Kaufmann_2023_ICCV}
Manuel Kaufmann, Jie Song, Chen Guo, Kaiyue Shen, Tianjian Jiang, Chengcheng Tang, Juan~Jos\'e Z\'arate, and Otmar Hilliges.
\newblock Emdb: The electromagnetic database of global 3d human pose and shape in the wild.
\newblock In \emph{Proceedings of the IEEE/CVF International Conference on Computer Vision (ICCV)}, pages 14632--14643, 2023.

\bibitem[Kocabas et~al.(2021)Kocabas, Huang, Hilliges, and Black]{pare}
Muhammed Kocabas, Chun-Hao~P Huang, Otmar Hilliges, and Michael~J Black.
\newblock Pare: Part attention regressor for 3d human body estimation.
\newblock In \emph{Proceedings of the IEEE/CVF International Conference on Computer Vision}, pages 11127--11137, 2021.

\bibitem[Kocabas et~al.(2024)Kocabas, Yuan, Molchanov, Guo, Black, Hilliges, Kautz, and Iqbal]{kocabas2024pace}
Muhammed Kocabas, Ye Yuan, Pavlo Molchanov, Yunrong Guo, Michael~J Black, Otmar Hilliges, Jan Kautz, and Umar Iqbal.
\newblock Pace: Human and camera motion estimation from in-the-wild videos.
\newblock In \emph{2024 International Conference on 3D Vision (3DV)}, pages 397--408. IEEE, 2024.

\bibitem[Kolotouros et~al.(2019)Kolotouros, Pavlakos, Black, and Daniilidis]{spin}
Nikos Kolotouros, Georgios Pavlakos, Michael~J Black, and Kostas Daniilidis.
\newblock Learning to reconstruct 3d human pose and shape via model-fitting in the loop.
\newblock In \emph{Proceedings of the IEEE/CVF international conference on computer vision}, pages 2252--2261, 2019.

\bibitem[Komura et~al.(2017)Komura, Habibie, Holden, Schwarz, and Yearsley]{komura2017recurrent}
Taku Komura, Ikhsanul Habibie, Daniel Holden, Jonathan Schwarz, and Joe Yearsley.
\newblock A recurrent variational autoencoder for human motion synthesis.
\newblock In \emph{The 28th British Machine Vision Conference}, 2017.

\bibitem[Laine et~al.(2020)Laine, Hellsten, Karras, Seol, Lehtinen, and Aila]{Laine2020diffrast}
Samuli Laine, Janne Hellsten, Tero Karras, Yeongho Seol, Jaakko Lehtinen, and Timo Aila.
\newblock Modular primitives for high-performance differentiable rendering.
\newblock \emph{ACM Transactions on Graphics}, 39\penalty0 (6), 2020.

\bibitem[Li et~al.(2021)Li, Villegas, Ceylan, Yang, Kuang, Li, and Zhao]{li2021task}
Jiaman Li, Ruben Villegas, Duygu Ceylan, Jimei Yang, Zhengfei Kuang, Hao Li, and Yajie Zhao.
\newblock Task-generic hierarchical human motion prior using vaes.
\newblock In \emph{2021 International Conference on 3D Vision (3DV)}, pages 771--781. IEEE, 2021.

\bibitem[Li et~al.(2002)Li, Wang, and Shum]{li2002motion}
Yan Li, Tianshu Wang, and Heung-Yeung Shum.
\newblock Motion texture: a two-level statistical model for character motion synthesis.
\newblock In \emph{Proceedings of the 29th annual conference on Computer graphics and interactive techniques}, pages 465--472, 2002.

\bibitem[Ling et~al.(2020)Ling, Zinno, Cheng, and Van De~Panne]{ling2020character}
Hung~Yu Ling, Fabio Zinno, George Cheng, and Michiel Van De~Panne.
\newblock Character controllers using motion vaes.
\newblock \emph{ACM Transactions on Graphics (TOG)}, 39\penalty0 (4):\penalty0 40--1, 2020.

\bibitem[Liu et~al.(2005)Liu, Hertzmann, and Popovi{\'c}]{liu2005learning}
C~Karen Liu, Aaron Hertzmann, and Zoran Popovi{\'c}.
\newblock Learning physics-based motion style with nonlinear inverse optimization.
\newblock \emph{ACM Transactions on Graphics (TOG)}, 24\penalty0 (3):\penalty0 1071--1081, 2005.

\bibitem[Liu et~al.(2021)Liu, Yang, Zhang, Cui, Rehg, and Tang]{egobody}
Miao Liu, Dexin Yang, Yan Zhang, Zhaopeng Cui, James~M Rehg, and Siyu Tang.
\newblock 4d human body capture from egocentric video via 3d scene grounding.
\newblock In \emph{2021 international conference on 3D vision (3DV)}, pages 930--939. IEEE, 2021.

\bibitem[Loper et~al.(2023{\natexlab{a}})Loper, Mahmood, Romero, Pons-Moll, and Black]{10.1145/3596711.3596800}
Matthew Loper, Naureen Mahmood, Javier Romero, Gerard Pons-Moll, and Michael~J. Black.
\newblock \emph{SMPL: A Skinned Multi-Person Linear Model}.
\newblock Association for Computing Machinery, New York, NY, USA, 1 edition, 2023{\natexlab{a}}.

\bibitem[Loper et~al.(2023{\natexlab{b}})Loper, Mahmood, Romero, Pons-Moll, and Black]{loper2023smpl}
Matthew Loper, Naureen Mahmood, Javier Romero, Gerard Pons-Moll, and Michael~J Black.
\newblock Smpl: A skinned multi-person linear model.
\newblock In \emph{Seminal Graphics Papers: Pushing the Boundaries, Volume 2}, pages 851--866. 2023{\natexlab{b}}.

\bibitem[Low et~al.(2020)Low, Gao, Xiang, and Ramesh]{low2020sofea}
Weng~Fei Low, Zhi Gao, Cheng Xiang, and Bharath Ramesh.
\newblock Sofea: A non-iterative and robust optical flow estimation algorithm for dynamic vision sensors.
\newblock In \emph{Proceedings of the IEEE/CVF Conference on Computer Vision and Pattern Recognition Workshops}, pages 82--83, 2020.

\bibitem[Mahmood et~al.(2019)Mahmood, Ghorbani, Troje, Pons-Moll, and Black]{mahmood2019amass}
Naureen Mahmood, Nima Ghorbani, Nikolaus~F Troje, Gerard Pons-Moll, and Michael~J Black.
\newblock Amass: Archive of motion capture as surface shapes.
\newblock In \emph{Proceedings of the IEEE/CVF international conference on computer vision}, pages 5442--5451, 2019.

\bibitem[Martinez et~al.(2017)Martinez, Hossain, Romero, and Little]{Martinez_2017_ICCV}
Julieta Martinez, Rayat Hossain, Javier Romero, and James~J. Little.
\newblock A simple yet effective baseline for 3d human pose estimation.
\newblock In \emph{Proceedings of the IEEE International Conference on Computer Vision (ICCV)}, 2017.

\bibitem[Moreno-Noguer(2017)]{Moreno-Noguer_2017_CVPR}
Francesc Moreno-Noguer.
\newblock 3d human pose estimation from a single image via distance matrix regression.
\newblock In \emph{Proceedings of the IEEE Conference on Computer Vision and Pattern Recognition (CVPR)}, 2017.

\bibitem[Pan et~al.()Pan, Liu, and Hartley]{pansingle}
Liyuan Pan, Miaomiao Liu, and Richard Hartley.
\newblock Single image optical flow estimation with an event camera. in 2020 ieee.
\newblock In \emph{CVF Conference on Computer Vision and Pattern Recognition (CVPR)}, pages 1669--1678.

\bibitem[Patel and Black(2024)]{camerahmr}
Priyanka Patel and Michael~J. Black.
\newblock Camerahmr: Aligning people with perspective, 2024.

\bibitem[Pavlakos et~al.(2017)Pavlakos, Zhou, Derpanis, and Daniilidis]{pavlakos2017coarse}
Georgios Pavlakos, Xiaowei Zhou, Konstantinos~G Derpanis, and Kostas Daniilidis.
\newblock Coarse-to-fine volumetric prediction for single-image 3d human pose.
\newblock In \emph{Proceedings of the IEEE conference on computer vision and pattern recognition}, pages 7025--7034, 2017.

\bibitem[Pavlakos et~al.(2018)Pavlakos, Zhou, and Daniilidis]{Pavlakos_2018_CVPR}
Georgios Pavlakos, Xiaowei Zhou, and Kostas Daniilidis.
\newblock Ordinal depth supervision for 3d human pose estimation.
\newblock In \emph{Proceedings of the IEEE Conference on Computer Vision and Pattern Recognition (CVPR)}, 2018.

\bibitem[Reda et~al.(2022)Reda, Kontkanen, Tabellion, Sun, Pantofaru, and Curless]{reda2022film}
Fitsum Reda, Janne Kontkanen, Eric Tabellion, Deqing Sun, Caroline Pantofaru, and Brian Curless.
\newblock Film: Frame interpolation for large motion.
\newblock In \emph{European Conference on Computer Vision}, pages 250--266. Springer, 2022.

\bibitem[Rempe et~al.(2021)Rempe, Birdal, Hertzmann, Yang, Sridhar, and Guibas]{Rempe_2021_ICCV}
Davis Rempe, Tolga Birdal, Aaron Hertzmann, Jimei Yang, Srinath Sridhar, and Leonidas~J. Guibas.
\newblock Humor: 3d human motion model for robust pose estimation.
\newblock In \emph{Proceedings of the IEEE/CVF International Conference on Computer Vision (ICCV)}, pages 11488--11499, 2021.

\bibitem[Rose et~al.(1998)Rose, Cohen, and Bodenheimer]{rose1998verbs}
Charles Rose, Michael~F Cohen, and Bobby Bodenheimer.
\newblock Verbs and adverbs: Multidimensional motion interpolation.
\newblock \emph{IEEE Computer Graphics and Applications}, 18\penalty0 (5):\penalty0 32--40, 1998.

\bibitem[Saini et~al.(2023)Saini, Huang, Black, and Ahmad]{smartmocap}
Nitin Saini, Chun-Hao~P Huang, Michael~J Black, and Aamir Ahmad.
\newblock Smartmocap: Joint estimation of human and camera motion using uncalibrated rgb cameras.
\newblock \emph{IEEE Robotics and Automation Letters}, 2023.

\bibitem[Scarpellini et~al.(2021)Scarpellini, Morerio, and Del~Bue]{scarpellini2021lifting}
Gianluca Scarpellini, Pietro Morerio, and Alessio Del~Bue.
\newblock Lifting monocular events to 3d human poses.
\newblock In \emph{Proceedings of the IEEE/CVF Conference on Computer Vision and Pattern Recognition}, pages 1358--1368, 2021.

\bibitem[Shao et~al.(2024)Shao, Wang, Zhou, Wang, Yang, and Li]{shao2024temporal}
Zhanpeng Shao, Xueping Wang, Wen Zhou, Wuzhen Wang, Jianyu Yang, and Youfu Li.
\newblock A temporal densely connected recurrent network for event-based human pose estimation.
\newblock \emph{Pattern Recognition}, 147:\penalty0 110048, 2024.

\bibitem[Stoffregen et~al.(2019)Stoffregen, Gallego, Drummond, Kleeman, and Scaramuzza]{stoffregen2019event}
Timo Stoffregen, Guillermo Gallego, Tom Drummond, Lindsay Kleeman, and Davide Scaramuzza.
\newblock Event-based motion segmentation by motion compensation.
\newblock In \emph{Proceedings of the IEEE/CVF International Conference on Computer Vision}, pages 7244--7253, 2019.

\bibitem[Sun et~al.(2017)Sun, Shang, Liang, and Wei]{Sun_2017_ICCV}
Xiao Sun, Jiaxiang Shang, Shuang Liang, and Yichen Wei.
\newblock Compositional human pose regression.
\newblock In \emph{Proceedings of the IEEE International Conference on Computer Vision (ICCV)}, 2017.

\bibitem[Sun et~al.(2022)Sun, Liu, Bao, Fu, Mei, and Black]{bev}
Yu Sun, Wu Liu, Qian Bao, Yili Fu, Tao Mei, and Michael~J Black.
\newblock Putting people in their place: Monocular regression of 3d people in depth.
\newblock In \emph{Proceedings of the IEEE/CVF Conference on Computer Vision and Pattern Recognition}, pages 13243--13252, 2022.

\bibitem[Tekin et~al.(2017)Tekin, Marquez-Neila, Salzmann, and Fua]{Tekin_2017_ICCV}
Bugra Tekin, Pablo Marquez-Neila, Mathieu Salzmann, and Pascal Fua.
\newblock Learning to fuse 2d and 3d image cues for monocular body pose estimation.
\newblock In \emph{Proceedings of the IEEE International Conference on Computer Vision (ICCV)}, 2017.

\bibitem[von Marcard et~al.(2018)von Marcard, Henschel, Black, Rosenhahn, and Pons-Moll]{Marcard_2018_ECCV}
Timo von Marcard, Roberto Henschel, Michael~J. Black, Bodo Rosenhahn, and Gerard Pons-Moll.
\newblock Recovering accurate 3d human pose in the wild using imus and a moving camera.
\newblock In \emph{Proceedings of the European Conference on Computer Vision (ECCV)}, 2018.

\bibitem[Wang et~al.(2024{\natexlab{a}})Wang, Wang, Liu, and Daniilidis]{wang2024tram}
Yufu Wang, Ziyun Wang, Lingjie Liu, and Kostas Daniilidis.
\newblock Tram: Global trajectory and motion of 3d humans from in-the-wild videos.
\newblock \emph{arXiv preprint arXiv:2403.17346}, 2024{\natexlab{a}}.

\bibitem[Wang et~al.(2022{\natexlab{a}})Wang, Chaney, and Daniilidis]{wang2022evac3d}
Ziyun Wang, Kenneth Chaney, and Kostas Daniilidis.
\newblock {EvAC3D}: From event-based apparent contours to {3D} models via continuous visual hulls.
\newblock In \emph{ECCV}, pages 284--299, 2022{\natexlab{a}}.

\bibitem[Wang et~al.(2022{\natexlab{b}})Wang, Cladera, Bisulco, Lee, Taylor, Daniilidis, Hsieh, Lee, and Isler]{wang2022ev}
Ziyun Wang, Fernando Cladera, Anthony Bisulco, Daewon Lee, Camillo~J Taylor, Kostas Daniilidis, M~Ani Hsieh, Daniel~D Lee, and Volkan Isler.
\newblock {EV-Catcher}: High-speed object catching using low-latency event-based neural networks.
\newblock 7\penalty0 (4):\penalty0 8737--8744, 2022{\natexlab{b}}.

\bibitem[Wang et~al.(2023)Wang, Hamann, Chaney, Jiang, Gallego, and Daniilidis]{wang2023event}
Ziyun Wang, Friedhelm Hamann, Kenneth Chaney, Wen Jiang, Guillermo Gallego, and Kostas Daniilidis.
\newblock Event-based continuous color video decompression from single frames.
\newblock \emph{arXiv preprint arXiv:2312.00113}, 2023.

\bibitem[Wang et~al.(2024{\natexlab{b}})Wang, Guo, and Daniilidis]{wang2025evimo}
Ziyun Wang, Jinyuan Guo, and Kostas Daniilidis.
\newblock Un-{EVIMO}: Unsupervised event-based independent motion segmentation.
\newblock In \emph{ECCV}, pages 228--245, 2024{\natexlab{b}}.

\bibitem[Xu et~al.(2020)Xu, Xu, Golyanik, Habermann, Fang, and Theobalt]{xu2020eventcap}
Lan Xu, Weipeng Xu, Vladislav Golyanik, Marc Habermann, Lu Fang, and Christian Theobalt.
\newblock Eventcap: Monocular 3d capture of high-speed human motions using an event camera.
\newblock In \emph{Proceedings of the IEEE/CVF Conference on Computer Vision and Pattern Recognition}, pages 4968--4978, 2020.

\bibitem[Ye et~al.(2018)Ye, Mitrokhin, Ferm{\"u}ller, Yorke, and Aloimonos]{ye2018unsupervised}
Chengxi Ye, Anton Mitrokhin, Cornelia Ferm{\"u}ller, James~A Yorke, and Yiannis Aloimonos.
\newblock Unsupervised learning of dense optical flow, depth and egomotion from sparse event data.
\newblock \emph{arXiv preprint arXiv:1809.08625}, 2018.

\bibitem[Ye et~al.(2023)Ye, Pavlakos, Malik, and Kanazawa]{slahmr}
Vickie Ye, Georgios Pavlakos, Jitendra Malik, and Angjoo Kanazawa.
\newblock Decoupling human and camera motion from videos in the wild.
\newblock In \emph{Proceedings of the IEEE/CVF Conference on Computer Vision and Pattern Recognition}, pages 21222--21232, 2023.

\bibitem[Zhang et~al.(2022)Zhang, Yezzi, and Gallego]{zhang2022formulating}
Zelin Zhang, Anthony~J Yezzi, and Guillermo Gallego.
\newblock Formulating event-based image reconstruction as a linear inverse problem with deep regularization using optical flow.
\newblock \emph{IEEE Transactions on Pattern Analysis and Machine Intelligence}, 45\penalty0 (7):\penalty0 8372--8389, 2022.

\bibitem[Zhou et~al.(2020)Zhou, Lu, Barnes, Yang, Xiang, et~al.]{zhou2020generative}
Yi Zhou, Jingwan Lu, Connelly Barnes, Jimei Yang, Sitao Xiang, et~al.
\newblock Generative tweening: Long-term inbetweening of 3d human motions.
\newblock \emph{arXiv preprint arXiv:2005.08891}, 2020.

\bibitem[Zhu et~al.(2018{\natexlab{a}})Zhu, Chen, and Daniilidis]{zhu2018realtime}
Alex~Zihao Zhu, Yibo Chen, and Kostas Daniilidis.
\newblock Realtime time synchronized event-based stereo.
\newblock In \emph{Proceedings of the European Conference on Computer Vision (ECCV)}, pages 433--447, 2018{\natexlab{a}}.

\bibitem[Zhu et~al.(2018{\natexlab{b}})Zhu, Yuan, Chaney, and Daniilidis]{zhu2018ev}
Alex~Zihao Zhu, Liangzhe Yuan, Kenneth Chaney, and Kostas Daniilidis.
\newblock Ev-flownet: Self-supervised optical flow estimation for event-based cameras.
\newblock \emph{arXiv preprint arXiv:1802.06898}, 2018{\natexlab{b}}.

\bibitem[Zhu et~al.(2019)Zhu, Yuan, Chaney, and Daniilidis]{zhu2019unsupervised}
Alex~Zihao Zhu, Liangzhe Yuan, Kenneth Chaney, and Kostas Daniilidis.
\newblock Unsupervised event-based learning of optical flow, depth, and egomotion.
\newblock In \emph{Proceedings of the IEEE/CVF Conference on Computer Vision and Pattern Recognition}, pages 989--997, 2019.

\bibitem[Zhu et~al.(2021)Zhu, Wang, Khant, and Daniilidis]{zhu2021eventgan}
Alex~Zihao Zhu, Ziyun Wang, Kaung Khant, and Kostas Daniilidis.
\newblock Eventgan: Leveraging large scale image datasets for event cameras.
\newblock In \emph{2021 IEEE international conference on computational photography (ICCP)}, pages 1--11. IEEE, 2021.

\bibitem[Zou et~al.(2021)Zou, Guo, Zuo, Wang, Wang, Hu, Chen, Gong, and Cheng]{zou2021eventhpe}
Shihao Zou, Chuan Guo, Xinxin Zuo, Sen Wang, Pengyu Wang, Xiaoqin Hu, Shoushun Chen, Minglun Gong, and Li Cheng.
\newblock Eventhpe: Event-based 3d human pose and shape estimation.
\newblock In \emph{Proceedings of the IEEE/CVF International Conference on Computer Vision}, pages 10996--11005, 2021.

\bibitem[Zou et~al.(2023)Zou, Mu, Zuo, Wang, and Cheng]{zou2023event}
Shihao Zou, Yuxuan Mu, Xinxin Zuo, Sen Wang, and Li Cheng.
\newblock Event-based human pose tracking by spiking spatiotemporal transformer.
\newblock \emph{arXiv preprint arXiv:2303.09681}, 2023.

\end{thebibliography}
}

\ifshowsupplementary
\clearpage
\setcounter{page}{1}
\maketitlesupplementary

\section{Implementation Details}
\subsection{Training Details}

\textbf{Multi-state Training.} We deploy a multi-stage training strategy, as briefly described in \cref{sec:method}. First, we train the Event Human Motion Predictor (E-HMP) described in \cref{sec:predictor} for 10 epochs. During this training, we use a dummy translation prediction network that directly regresses the poses as in~\cite{zou2021eventhpe}, which is then later removed. In parallel, we train the Glboal Motion Predictor (GMP) described in \cref{sec:adaptation} for 5 epochs. At this point, the GMP network is slightly overfitted to the training data. We then freeze the GMP and merge with the E-HMP to help convergence, and then unfreeze it after 1 epoch and decrease the learning rate to 1e-5 to jointly fine-tune the two networks. Empirically, we find that this iterative training approach helps convergence for both the GMP and the E-HMP. For each component of the multi-stage training, we use a start learning rate of 1e-3 and then decay it to 1e-4 after one epoch to help stabilize training. We use batch size of 16 for all networks.

\textbf{Hyperparameters.} In all experiments, we set $\lambda_{ori} = 10$, $\lambda_t = 10$, $\lambda_{3D} = 20$, $\lambda_{2D} = 20$, $\lambda_{flow} = 0.1$ and $\lambda_c = 0.1$. The starting learning rate is 1e-3 for all networks trained from scratch and decayed by 10 after the first epoch. We use a learning rate of 1e-5 for fine-tuning the GMP network. We use Adam as optimizer for all experiments.

\textbf{Global Motion Field versus Global Motion Predictor} Here, we disambiguate between the two types of ``global" motions mentioned in the main paper. The global motion field, which is decoded from the latent code $z_{g}$, represents the relative rotation of the root joint. These rotations are important to convert joint positions into the root-adjusted global pose, which are needed for the Global Motion Predictor. On the other hand, the Global Motion Predictor (GMP), predicts the root velocity based on the input joint positions, joint velocities, joint orientations, and rotational velocities. The GMP does not use $z_g$ directly. Instead, it uses pelvis-centered local poses (decoded from $z_l$) rotated by each root orientation (decoded from $z_g$) to compute the translational velocities estimate.

\section{Architecture}
\subsection{Event Human Motion Predictor (E-HMP)}
First, we use a pre-trained ResNet50~\claude{citation} image encoder pre-trained on ImageNet~\claude{citation} to compute features from the event volumes. The first convolutions layers with three input channels are replaced with ones with eight input channels. The output features at different steps are processed with a Gated Recurrent Unit network that recurrently updates a hidden state. In the last step $t_T$, we use a linear layer to project the hidden state to the a vector of size $1024 + 256$, which is then divided into the global code $z_g$ and $z_l$. The GRU has one layer with $2048$ hidden size with no dropout and with one direction. \textbf{Multi-layer Preceptron Motion Decoder} Our continuous decoding rely on an MLP that takes timestamps as input and outputs the pose parameters. We used an MLP with 11 linear layers with ReLU nonlinearities and skip connections. The input timestamps are mapped to positional encoding and concatenated with the latent codes, as input to the MLP.

\subsection{Global Motion Predictor (GMP)}
Following~\cite{aberman2020skeleton} \claude{fix this}, we use Skeleton Convolutions to aggregate features on the graph defined by the SMPL skeleton. The local poses parameters pass through 3 layers of convolution layers, followed by four more 1D convolution layers and average pooling layers to obtain the final translational velocities. Each pose predicts on velocity vector. No temporal blending is done at this step.

\section{Additional Human Mesh Prediction Results}
In~\cref{supp:fig:sequence}, we show the results of human predictions at five different frame rates. Due to the continuous nature of our Event Human Motion Predictor (E-HMP), the poses can be queried at any timestamp during the event duration. We only show 120 FPS here because it is difficult to visualize a higher frame rate in a 2D image. Animated results can be found in the supplementary video. Each row corresponds to a 1 second sequence of human motion. We overlaid the predictions at different timestamps onto the same image. It can be observed that our network allows 120 FPS decoding in parallel, which produces smooth and continuous-time movements. We note that the continuous human motion field is predicted at once from the events, and sampling poses at different timestamps requires only inputting arbitrary timestamps to the lightweight MLP.
\setcounter{section}{0}
\section{\ourdata{} (\ourdatashort{})}

\subsection{Collection setup}
\ourdatashort{} uses four high-speed Flir HD RGB cameras to collect images for annotations, will be referred to as GT cameras. And a beam splitter with a Flir RGB camera(2448 $\times$ 2048) and a Prophesee Event camera(VGA) to capture paired event data and images. The beam splitter and trigger box design are shown in \cref{fig:beam}. The setup of the cameras is shown in \cref{fig:setup}. 

\begin{figure}[ht]
    \centering
    \includegraphics[clip, trim={8.2cm, 7.5cm, 8cm, 8cm}, width=\linewidth]{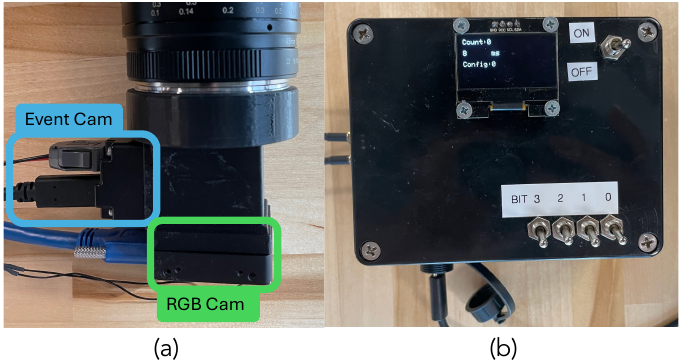}
    \caption{\label{fig:beam}\textbf{Beam splitter and trigger box design} \textbf{a} Beam splitter with an event camera and a paired RGB camera. \textbf{b} Trigger box design with adjustable time interval, integrating an Arduino Teensy board.}
\end{figure}

\begin{figure}[ht]
    \centering
    \includegraphics[clip, trim={9cm, 7cm, 8.5cm, 7cm}, width=\linewidth]{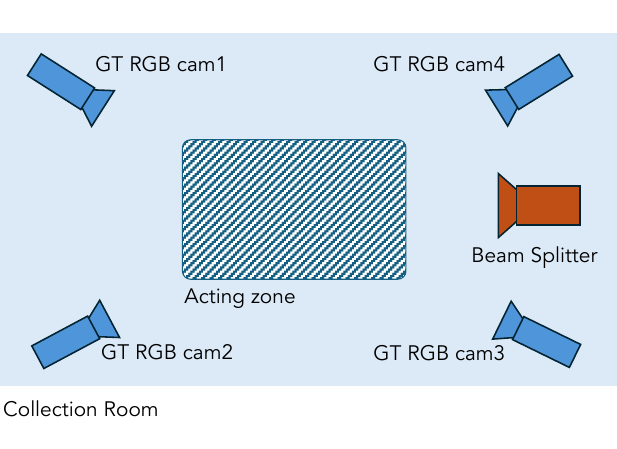}
    \includegraphics[clip, trim={8.5cm, 7cm, 8.5cm, 7cm}, width=\linewidth]{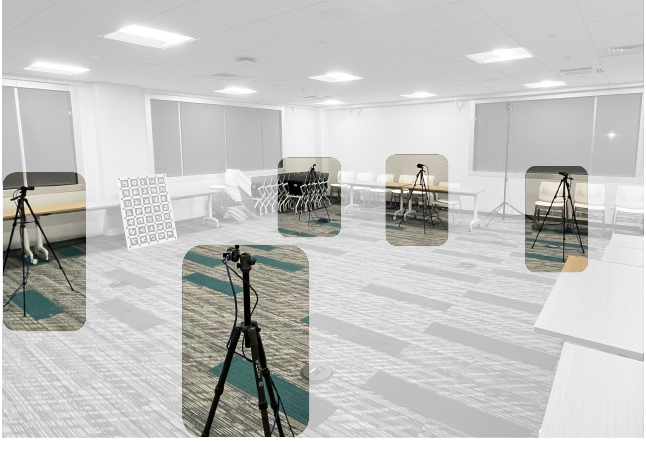}
    \caption{\label{fig:setup}\textbf{\ourdatashort{} collection setup.} 4 high speed cameras for gt are placed at 4 corners and a beam splitter is placed at one side to capture human in the middle of the room. The space of acting zone is 1.5m $\times$ 2m. }
\end{figure}

Before data collection, all cameras, including the event camera, are calibrated using Kalibr~\cite{}. The calibration procedure for the event camera follows the methodology in~\cite{}. The reprojection error for each camera is less than 1.0 pixel. Both the calibration results and the collected sequences are publicly released as part of \ourdatashort{}.

During collections, gt cameras are triggered at 125 Hz with exposure time set to 5 ms to minimize motion blur. The RGB camera within the beam splitter is triggered at 15.625 Hz (division of 8) with an exposure time of 50 ms to produce intentionally blurry frames for comparison. Event camera is triggered with the same rate as gt cameras. We evaluate the trigger drift of our 125 Hz synchronizing signal. As is shown in \cref{fig:bias}, the trigger interval has a minimal drift of on average 0.01938 \textperthousand. All RGB cameras are set to use auto-gain.

\begin{figure}[ht]
    \centering
    \includegraphics[clip, trim={0cm, 0cm, 0cm, 0cm}, width=1.\linewidth]{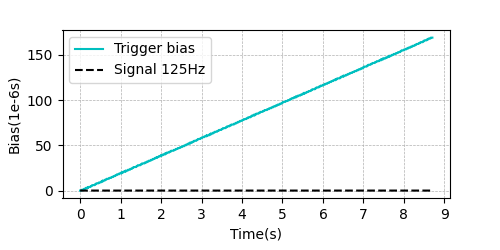}
    \caption{\label{fig:bias}\textbf{Hardware synchronization signal evaluation} (a) We evaluated the bias drift of triggers by recording the timestamp of the trigger subtracted by a standard 125 Hz signal, the total drift over 8.6 s is 0.169 ms.}
    \vspace{-.5cm}
\end{figure}

\begin{table*}[ht]
\centering
\caption{\label{tab:dataset}\ourdatashort{} dataset details. \ourdatashort{} consists of 40 different human motions that are categorized into \textbf{Basic}, \textbf{Medium} and \textbf{Extreme} levels. \textbf{Basic} activities include motion of arms, legs, main body and head. \textbf{Medium} activities are common actions involving more than 2 parts of the human body. \textbf{Extreme} activities are competitive sports with fast motions}
\begin{tabular}{l|ccccc}
\toprule
\textbf{Category} & & &\textbf{Actions} & & \\
\midrule
\multicolumn{6}{c}{\textbf{Basic}}\\
\midrule
Arms        & Left arm upward    & Right arm upward & Bicep curl & Left arm wave & Right arm wave   \\
            & Left arm circle       & Right arm circle     & Left punch & Right punch & Punch\\
            & Left arm raise       & Right arm raise     & Left arm outward     & Right arm outward & Outward  \\
Legs        & Left knee lift        & Right knee lift      & Left hop         & Right hop & Left kick \\
            & Right kick  & Left lunge & Right lunge & \\
Body & Lean left      & Lean right          & Rotate torso          & Rotate head & Nod \\
\midrule
\multicolumn{6}{c}{\textbf{Medium}}\\
\midrule
Common &  Walk & Jog & Jump up-down & Jump forward-back & Jump sideway\\
 Activities   & Starjump & Squat & & & \\
\midrule
\multicolumn{6}{c}{\textbf{Extreme}}\\
\midrule
Sports & Taekwondo & Tennis & Volleyball & Gymnastics & Shot put \\
\bottomrule
\end{tabular}
\end{table*}

Using four views from different directions, data in \ourdatashort{} is annotated with Easymocap~\cite{easymocap}, which employs 2D joint detection and triangulation to determine 3D joint locations. Examples of the annotation results are shown in Fig.~\ref{supp:fig:anno}. In the figure, the annotated SMPL model is overlaid on the original images to visually assess annotation quality.

In our collections 9 out of 160 sequences are discarded due to inaccurate annotations, primarily caused by occlusions of certain parts of the human body. 

\subsection{Sequences}
\claude{Rex}
In total, \ourdatashort{} consists of 160 sequences with 40 different motions and 200 thousand frames of SMPL annotations. The motion are categorized by prediction difficulty into 3 increasing levels, varying from basic motion such as arm abduction to extreme sports such as Taekwondo, Volleyball and Tennis. The details of captured motions are described in \cref{tab:dataset}

We present 5 example sequences: Left arm wave, Bicep curl, Left lunge, Lean left, Volleyball of our \ourdatashort{} dataset in Fig.\ref{supp:fig:anno}.

\begin{figure*}[h]
    \centering
    \includegraphics[clip, trim={0cm, 1.5cm, 0cm, 1.5cm}, width=\linewidth]{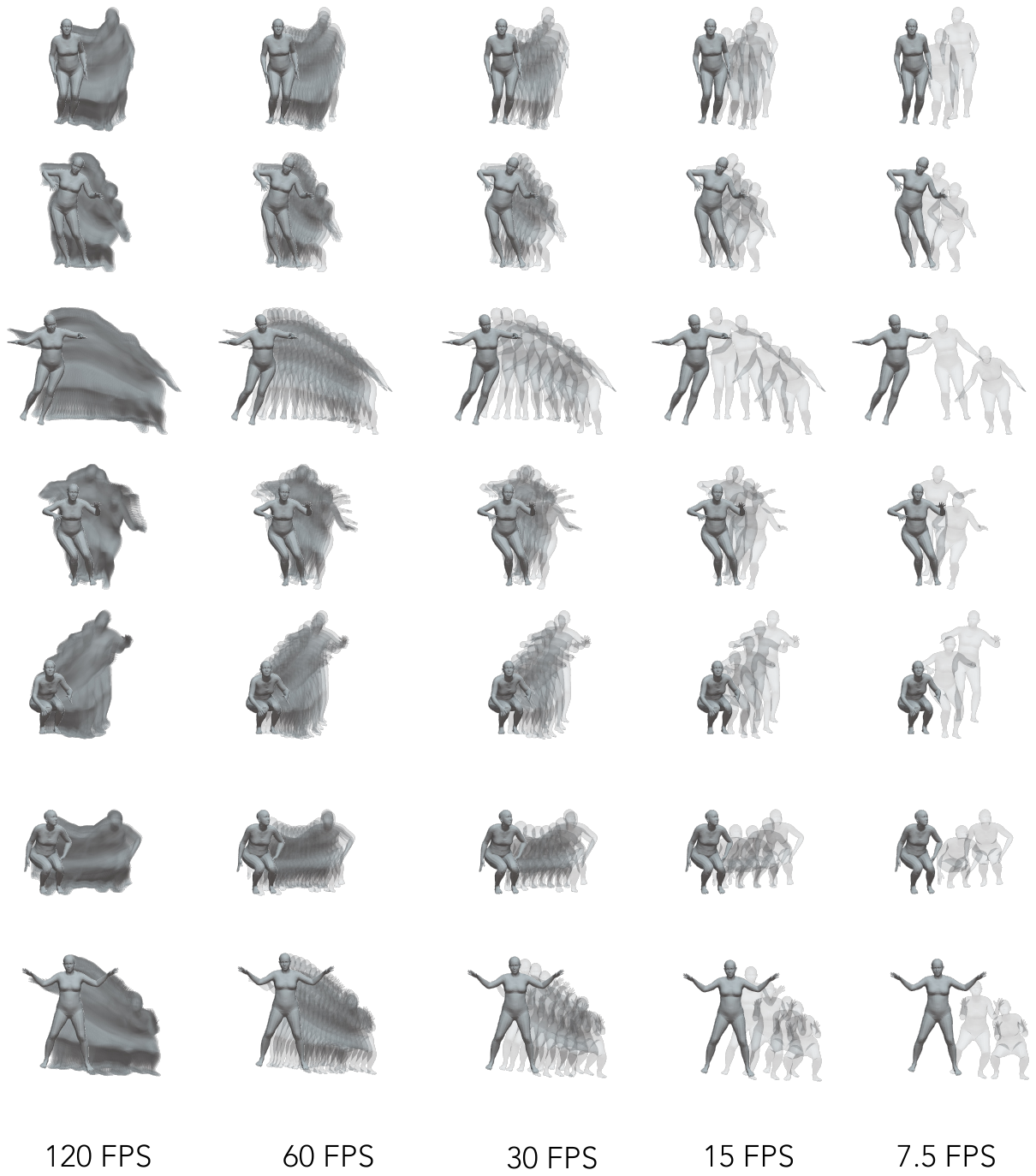}
    \caption{Predicted Human Motion Sequence. We show the sequences of human motion in each image by overlaying predictions at different timestamps. Past predictions are rendered with high transparency.}
    \label{supp:fig:sequence}
\end{figure*}

\begin{figure*}[h]
    \centering
    \includegraphics[clip, trim={7.5cm, 5.6cm, 10.4cm, 4.2cm}, width=\linewidth]{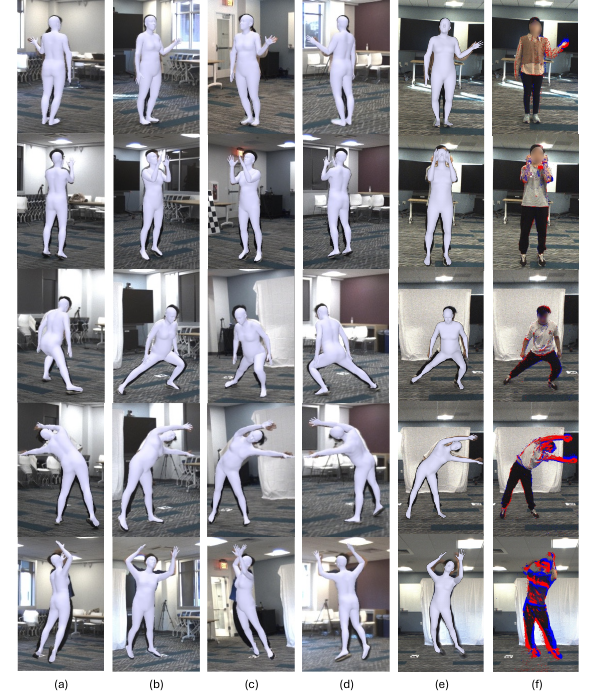}
    \caption{We present six example sequences from \ourdatashort{}. Each sequence, from left to right, includes: \textbf{(a-d)} Four multi-camera images with SMPL annotation overlaid via EasyMocap~\cite{easymocap}. \textbf{(e)} The estimated mesh model superimposed on the beam splitter RGB camera. \textbf{(f)} Events displayed on the beam splitter RGB camera. }
    \label{supp:fig:anno}
\end{figure*}

\section{Baseline Methods}

\subsection{DHP19}
\claude{Rex}
Since the annotation method differs between \ourdatashort{} and DHP19 data, we re-trained the DHP19 network on our dataset using 24 joints derived from the annotated SMPL model. To generate ground truth for the DHP19 network, SMPL annotations are converted and projected into 24 2D joint positions, which are then rearranged into a heatmap with 24 channels, each representing the probability of one joint. The output dimension of the DHP19 network is adjusted to 24 accordingly. For smoothing the heatmaps, we use a 15-pixel Gaussian blur kernel from Torchvision with sigma of 2.6 pixels for computational efficiency. The whole implementation is built using PyTorch and additional details such as data format, optimizer, and learning rate decay strategy follow the setup of the original paper.

\subsection{EventCap}
\claude{Zi-Yan}
EventCap \cite{xu2020eventcap} is a groundbreaking approach for capturing high-speed human motions using a single event camera, delivering high-quality motion results at an impressive 1000 fps.  
We followed the steps outlined in the original EventCap paper\cite{xu2020eventcap}  with modifications to adapt the algorithm to both the MMHPSD dataset and our \ourdatashort{} dataset. We adapted the EventCap baseline to use SMPL models instead of the proprietary scanned textured human models in the original implementation. We initialized SMPL parameters using HMR 2.0 outputs through slerp interpolation.

The entire EventCap\cite{xu2020eventcap} pipeline is divided into two main components: batch optimization and event refinement. In the batch optimization stage, we included four loss terms: correspondence loss, 2D loss, 3D loss, and temporal loss. For feature trajectory, we experimented with both event feature tracking and image feature tracking. However, we found that image feature tracking provided more stable performance. To give EventCap a fair shot, we report the better performance using the image-based features. In the temporal loss term, we assessed the distance of events surrounding specific joints to determine their inclusion in the energy function, thereby introducing a temporal stabilization constraint for non-moving body parts. Since the reimplementation in \claude{EventHPE\cite{zou2021eventhpe}} did not specify the weight of each term, we empirically determine the weight for each term to prevent optimization collapse caused by poorly initialized joints being misclassified as non-moving.
The chosen hyperparameters chosen are
\begin{align}
\lambda_{3D}= 10,
\lambda_{2D}= 10,
\lambda_{temp} = 0.01,
\lambda_{cor} = 2.5.
\end{align}
The event refinement implementation adhered largely to the original paper\cite{xu2020eventcap},with some parameter adjustments for improved performance. We collect the closest events corresponding to the silhouette' 8x8 neighborhood and build the criterion for ICP with $\lambda_{dist} = 0.5$.  \claude{too vague. what are the changes?} The entire EventCap pipeline was implemented in PyTorch with SGD as the optimizer.

\fi

\end{document}